\documentclass[11pt]{article}          
\PassOptionsToPackage{table}{xcolor}
\usepackage[margin=1in]{geometry}      

\usepackage[utf8]{inputenc}
\usepackage[T1]{fontenc}
\usepackage{lmodern}                   
\usepackage{booktabs}
\usepackage{longtable}
\usepackage{amsmath,amssymb}
\usepackage{graphicx}
\usepackage{wrapfig}
\usepackage{tikz}
\usepackage{pgfplots}
\pgfplotsset{compat=1.18}
\usetikzlibrary{arrows.meta,positioning,calc,fit,backgrounds}
\usepackage{xcolor}

\definecolor{cRep}{HTML}{0B6FA4}
\definecolor{cPara}{HTML}{D2691E}
\definecolor{slate}{HTML}{3C4858}
\definecolor{cGood}{HTML}{2E8B57}
\definecolor{cBad}{HTML}{C0392B}
\definecolor{JSlightg}{HTML}{EAF7EE}
\definecolor{JSlightr}{HTML}{FCE8E8}
\definecolor{JSlightorange}{HTML}{FFF0E6}
\definecolor{JSlightblue}{HTML}{EAF4FB}
\definecolor{JSgrey}{HTML}{666666}
\definecolor{JSgreen}{HTML}{2E8B57}
\definecolor{JSred}{HTML}{C43C35}
\definecolor{JSorange}{HTML}{E66100}
\definecolor{JSblue}{HTML}{0072B2}
\definecolor{repBlue}{HTML}{1F6FB2}
\definecolor{parOrange}{HTML}{E2731B}
\definecolor{inkGrey}{HTML}{333A45}
\definecolor{okGreen}{HTML}{2E9E63}
\definecolor{endPink}{HTML}{E0555F}
\definecolor{vAlibaba}{HTML}{0072B2}
\definecolor{vMeta}{HTML}{D55E00}
\definecolor{vAnthropic}{HTML}{CC79A7}
\definecolor{vDeepSeek}{HTML}{009E73}
\definecolor{vGoogle}{HTML}{E69F00}
\definecolor{vMistral}{HTML}{56B4E9}
\definecolor{vZhipu}{HTML}{8C6D31}
\definecolor{vMoonshot}{HTML}{555555}
\definecolor{gridGrey}{HTML}{D8D8D8}
\definecolor{ruleGrey}{HTML}{8A8A8A}
\definecolor{cbBlue}{HTML}{0072B2}
\definecolor{cbOrange}{HTML}{E69F00}
\definecolor{cbGreen}{HTML}{009E73}
\definecolor{cbPurple}{HTML}{CC79A7}
\definecolor{cbRed}{HTML}{D55E00}
\definecolor{cbBrown}{HTML}{8C6D31}
\usepackage{colortbl}
\usepackage{etoolbox}
\AtBeginEnvironment{tabular}{}
\AtBeginEnvironment{longtable}{}
\definecolor{headBand}{HTML}{E9EFF5}
\definecolor{judgeBand}{HTML}{F5F8FB}
\definecolor{chipGrey}{HTML}{9AA5B1}

\newcommand{\dvis}[1]{%
  \ifdim #1pt>0pt\relax\colorlet{dvc}{okGreen}\else
    \ifdim #1pt>-0.02pt\relax\colorlet{dvc}{chipGrey}\else
      \ifdim #1pt>-0.10pt\relax\colorlet{dvc}{parOrange}\else
        \colorlet{dvc}{cBad}\fi\fi\fi
  \begin{tikzpicture}[baseline=-0.62ex,x=1cm,y=1cm]
    \useasboundingbox (-1.30,-0.115) rectangle (0.16,0.115);
    \fill[dvc] (0,-0.052) rectangle ({#1*3.7},0.052);
    \draw[cBad!55,line width=0.35pt] (-0.074,-0.105) -- (-0.074,0.105);
    \draw[black!45,line width=0.5pt] (0,-0.105) -- (0,0.105);
  \end{tikzpicture}}
\newcommand{\dnum}[1]{%
  \ifdim #1pt>0pt\relax\textcolor{okGreen}{$#1$}\else
    \ifdim #1pt>-0.02pt\relax\textcolor{chipGrey}{$#1$}\else
      \ifdim #1pt>-0.10pt\relax\textcolor{parOrange}{$#1$}\else
        \textcolor{cBad}{$#1$}\fi\fi\fi}
\usepackage{microtype}
\usepackage[numbers,round]{natbib}
\usepackage[hidelinks]{hyperref}
\usepackage{url}

\title{JudgeSense: A Benchmark for Prompt Sensitivity in LLM-as-a-Judge Systems}

\newif\ifblind
\blindfalse                             
\ifdefined\blindsubmission\blindtrue\fi 

\usepackage{orcidlink}
\newcommand{\orcidRohith}{}
\newcommand{\orcidWenbin}{}
\newcommand{\showorcid}[1]{\ifx\relax#1\relax\else\,\orcidlink{#1}\fi}

\newcommand{\blfootnote}[1]{%
  \begingroup\renewcommand{\thefootnote}{}\footnote{#1}%
  \addtocounter{footnote}{-1}\endgroup}

\ifblind
  \author{Anonymous authors\\Paper under double-blind review}
\else
  \author{
    Rohith Reddy Bellibatlu\showorcid{\orcidRohith}$^{*1}$ \quad
    Edward Raff$^{2,3}$ \quad
    Wenbin Zhang\showorcid{\orcidWenbin}$^{1}$ \\[0.6em]
    {\normalsize $^{1}$Florida International University, Miami, FL} \\
    {\normalsize $^{2}$CrowdStrike, New York, USA} \\
    {\normalsize $^{3}$University of Maryland Baltimore County, Baltimore,
     USA} }
\fi
\date{}

\begin{document}
\maketitle
\ifblind\else
\blfootnote{$^{*}$Corresponding author.}
\blfootnote{Benchmark and decision logs:
  \url{https://huggingface.co/datasets/Rohithreddybc/judgesense-benchmark}.
  Code, and the scripts behind every table and figure:
  \url{https://github.com/rohithreddybc/judgeSense}.}
\fi

\begin{abstract}
Large language models are widely used to judge the output of other language
models, yet whether a judge returns the same verdict when the same request is
worded differently remains largely unexamined. We study that question across
four evaluation tasks and twenty-five judges from six providers. To support the
analysis we release \textsc{JudgeSense}, a benchmark of $880$ items from
human-labelled corpora, each issued under two instructions that differ in
wording and not in what they ask, with the complete decision logs. Every score
is reported against the judge's own agreement with itself on the identical
prompt, so decoding noise is not charged to wording, and the release lets a
reader ask the same of any judge not in our roster. Rewording costs agreement on
all four tasks, and on two it clears the threshold we declare for a practically
meaningful effect; the ordinal task is both the least stable and the one fewest
judges are accurate on, and within a single family parameter count does not
predict stability. A judge measured inside an agent harness yields a
\emph{smaller} estimate than the same judge reached through a direct API call,
because its agreement with itself collapses faster than its agreement across
wordings.
\end{abstract}

\section{Introduction}
\label{sec:intro}

Language models now routinely evaluate the output of other language models.
A model used this way is a \emph{judge}: it is shown one piece of text, or two,
together with an instruction telling it what to decide, and it returns a label
from a small fixed set. Throughout this paper ``judge'' means the model doing
the deciding, never the model whose output is being judged. The practice is
cheap, fast, and correlates well enough with human preference to have displaced
human annotation across much of the evaluation stack
\citep{zheng2023judging}, and an evaluator earns that role by being
reproducible.
If two researchers ask the same judge the same question in different words and
get different verdicts, the judge is partly measuring the phrasing. The usual
validation cannot see that: it asks whether a judge agrees with humans on
average, and a judge can correlate well in aggregate while flipping on
individual items.

Instability under rephrasing is well documented for models performing ordinary
tasks, where formatting, ordering and the choice between synonyms move accuracy
substantially \citep{sclar2024quantifying, mizrahi2024state,
elazar2021measuring}. For judges the known failures are positional and
length-related: judges favour the first-presented candidate
\citep{wang2023fair, shi2024judging}, and annotators in the preference votes we
draw on chose the longer response in roughly 70\% of usable comparisons.
Whether the same judge, asked the same question two ways, returns the same
answer has been measured far less; Section~\ref{sec:related} sets out what has
and has not been established.

This paper measures it. We introduce JudgeSense, 880 items across four
evaluation tasks, each issued under two instruction templates that differ in
wording and not in what they ask. \emph{Factuality} asks whether a statement is
factually correct and \emph{coherence} asks how well a summary holds together
on a one-to-five scale; \emph{relevance} asks which of two passages better
answers a query and \emph{preference} which of two responses is better. The
four are independent task families with their own items and their own
paraphrase sets, and no item appears in more than one. The quantity we report is the Judge
Sensitivity Score,
\begin{equation}
\label{eq:jss}
\mathrm{JSS}(j, t) \;=\; \frac{1}{|P_t|} \sum_{(p, p') \in P_t}
  \delta\!\left( j(p),\, j(p') \right),
\end{equation}
where $j$ is a judge, $t$ an evaluation task, and $P_t$ that task's set of
\emph{prompt-paraphrase pairs}. $\delta(a,b) = 1$ if $a = b$ and $0$ otherwise,
evaluated in the task's decision space $D_t$. A JSS of $1.0$ means the judge
returned identical decisions on every pair; below $0.8$, rewording alone flips
its verdict on more than one item in five. This is not accuracy. A judge that
is wrong on every item but wrong \emph{identically} under both phrasings scores
$1.0$, which is intended: we ask whether the instrument is stable and treat
whether it is correct as a separate question. One judge measured on one task is
a \emph{cell}, the unit every table and figure below is built from.

A prompt-paraphrase pair is two evaluation instructions that ask for the same
decision on the same item and differ only in surface wording. One factuality
pair, shipped as templates T1 and T3, reads in full:
\begin{quote}\ttfamily\small\raggedright
Is this statement factually correct? Answer YES or NO only. \{text\}\\[2pt]
Is the following statement accurate? Answer with one word: YES or NO. \{text\}
\end{quote}
Both request the same two labels in the same direction, the statement
substituted into each is byte-identical, and neither asks the negated question.
The paraphrase is of the \emph{instruction}, never of the item it carries.

\begin{figure}[!t]
\centering
\resizebox{0.78\linewidth}{!}{%
\begin{tikzpicture}[x=1cm,y=1cm,font=\sffamily\small,>=Latex]
\tikzset{
 box/.style={draw=black!55,rounded corners=1.5pt,minimum height=.72cm,align=center,inner xsep=5pt,fill=white},
 bblue/.style={box,draw=JSblue,fill=JSlightblue},
 borange/.style={box,draw=JSorange,fill=JSlightorange},
 judge/.style={box,fill=black!7,minimum width=1.25cm,minimum height=1.55cm},
 verdictb/.style={bblue,minimum width=.95cm},
 verdicto/.style={borange,minimum width=.95cm},
 arrow/.style={->,line width=.65pt,draw=black!55},
 bluearrow/.style={->,line width=.8pt,draw=JSblue},
 orangearrow/.style={->,line width=.8pt,draw=JSorange}
}

\node[font=\bfseries] at (-.1,4.65) {(a)};
\node[font=\bfseries,anchor=west] at (.25,4.65) {One item is judged four times by the same model};
\node[box,minimum width=1.35cm,minimum height=1.15cm] (item) at (.10,2.95) {one\\item};

\node[bblue,minimum width=3.25cm] (A) at (3.95,4.00) {template A};
\node[bblue,dashed,minimum width=3.25cm] (Ap) at (3.95,3.10) {template A \emph{again}};
\node[borange,minimum width=3.25cm] (B) at (3.95,2.15) {template B (reworded)};
\node[borange,dashed,minimum width=3.25cm] (Bp) at (3.95,1.25) {template B \emph{again}};

\node[judge] (j) at (7.30,2.62) {the\\judge};
\node[verdictb] (dA) at (9.00,4.00) {$d_A$};
\node[verdictb] (dAp) at (9.00,3.10) {$d_{A'}$};
\node[verdicto] (dB) at (9.00,2.15) {$d_B$};
\node[verdicto] (dBp) at (9.00,1.25) {$d_{B'}$};

\foreach \n/\dy in {A/0.30, Ap/0.10, B/-0.10, Bp/-0.30}{%
  \draw[arrow] ([yshift=\dy cm]item.east) -- (\n.west);}
\foreach \n/\y in {A/3.15, Ap/2.85, B/2.45, Bp/2.15}{%
  \draw[arrow] (\n.east) -- (j.west|-0,\y);}
\foreach \n/\y in {dA/3.15, dAp/2.85, dB/2.45, dBp/2.15}{%
  \draw[arrow] (j.east|-0,\y) -- (\n.west);}

\draw[JSblue,line width=1pt] (9.73,4.22) -- (9.98,4.22) -- (9.98,2.88) -- (9.73,2.88);
\node[anchor=west,text=JSblue,font=\bfseries] at (10.10,3.55) {$\mathrm{JSS}_{\text{rep}}$};
\node[anchor=west,text=JSgrey,font=\itshape\scriptsize,align=left] at (10.10,3.10) {same words twice:\\does it agree with itself?};

\draw[JSorange,line width=1pt] (9.73,2.37) -- (9.98,2.37) -- (9.98,1.03) -- (9.73,1.03);
\node[anchor=west,text=JSorange,font=\bfseries] at (10.10,1.70) {$\mathrm{JSS}_{\text{para}}$};
\node[anchor=west,text=JSgrey,font=\itshape\scriptsize,align=left] at (10.10,1.24) {reworded:\\does the verdict survive?};

\node[fill=JSlightr,rounded corners=2pt,inner xsep=8pt,inner ysep=5pt,font=\bfseries] at (5.05,.38) {$\Delta\mathrm{JSS}=\textcolor{JSorange}{\mathrm{JSS}_{\text{para}}}-\textcolor{JSblue}{\mathrm{JSS}_{\text{rep}}}$};
\node[text=JSgrey,font=\itshape\scriptsize] at (5.05,-.10) {what rewording costs, once the judge's own noise is taken out};

\node[font=\bfseries] at (-.1,-1.00) {(b)};
\node[font=\bfseries,anchor=west] at (.25,-1.00) {A low ceiling hides wording sensitivity};

\path[fill=JSlightg,rounded corners=3pt] (.15,-4.05) rectangle (4.65,-1.35);
\node[text=JSgreen,font=\bfseries\scriptsize,align=center] at (2.40,-1.76) {High repeat reliability\\(readable)};
\node[font=\scriptsize] at (2.40,-2.28) {large gap (e.g., $-0.15$)};
\draw[->,draw=black!45] (.55,-3.38) -- (4.25,-3.38);
\foreach \x/\lab in {.6/0.6,.7/0.7,.8/0.8,.9/0.9,1/1.0}{
  \draw[black!45] ({.55+(\x-.6)*9.25},-3.32) -- ++(0,-.12);
  \node[font=\scriptsize] at ({.55+(\x-.6)*9.25},-3.68) {\lab};
}
\coordinate (lp) at ({.55+(.71-.6)*9.25},-2.95);
\coordinate (lr) at ({.55+(.90-.6)*9.25},-2.95);
\draw[line width=4pt,draw=JSorange!35] (lp) -- (lr);
\node[draw=JSorange,fill=white,minimum size=6pt,inner sep=1pt] at (lp) {};
\fill[JSblue] (lr) circle (3.2pt);

\path[fill=JSlightr,rounded corners=3pt] (4.95,-4.05) rectangle (9.45,-1.35);
\node[text=JSred,font=\bfseries\scriptsize,align=center] at (7.20,-1.76) {Low repeat reliability\\(not read)};
\node[font=\scriptsize] at (7.20,-2.28) {small gap (e.g., $-0.03$)};
\draw[->,draw=black!45] (5.35,-3.38) -- (9.05,-3.38);
\foreach \x/\lab in {.6/0.6,.7/0.7,.8/0.8,.9/0.9,1/1.0}{
  \draw[black!45] ({5.35+(\x-.6)*9.25},-3.32) -- ++(0,-.12);
  \node[font=\scriptsize] at ({5.35+(\x-.6)*9.25},-3.68) {\lab};
}
\coordinate (rp) at ({5.35+(.83-.6)*9.25},-2.95);
\coordinate (rr) at ({5.35+(.86-.6)*9.25},-2.95);
\draw[line width=4pt,draw=JSorange!35] (rp) -- (rr);
\node[draw=JSorange,fill=white,minimum size=6pt,inner sep=1pt] at (rp) {};
\fill[JSblue] (rr) circle (3.2pt);
\draw[dashed,JSred,line width=.8pt] ({5.35+(.9-.6)*9.25},-2.50) -- ({5.35+(.9-.6)*9.25},-3.45);

\node[draw=JSorange,fill=white,minimum size=6pt,inner sep=1pt] at (.55,-4.47) {};
\node[anchor=west,font=\scriptsize] at (.72,-4.47) {paraphrase (reworded)};
\fill[JSblue] (4.05,-4.47) circle (3.2pt);
\node[anchor=west,font=\scriptsize] at (4.22,-4.47) {repeat (identical)};
\draw[dashed,JSred] (6.80,-4.47) -- (7.25,-4.47);
\node[anchor=west,font=\scriptsize] at (7.42,-4.47) {repeat ceiling $=0.90$};

\node[font=\bfseries] at (-.1,-5.05) {(c)};
\node[font=\bfseries,anchor=west] at (.25,-5.05) {What rewording costs, pooled over judges and clustered by vendor};

\path[fill=black!4,rounded corners=3pt] (.15,-7.45) rectangle (9.45,-5.40);
\draw[JSred,dashed,line width=.7pt] (5.75,-5.52) -- (5.75,-7.33);
\draw[black!50,line width=.6pt] (6.35,-5.52) -- (6.35,-7.33);

\fill[JSred] (6.35,-5.78) rectangle (0.92,-5.62);
\node[anchor=east,font=\scriptsize] at (0.80,-5.70) {$-0.181$};
\node[anchor=west,font=\scriptsize] at (6.48,-5.70) {coherence};

\fill[JSred] (6.35,-6.28) rectangle (4.64,-6.12);
\node[anchor=east,font=\scriptsize] at (4.52,-6.20) {$-0.057$};
\node[anchor=west,font=\scriptsize] at (6.48,-6.20) {factuality};

\fill[JSred!28] (6.35,-6.78) rectangle (5.00,-6.62);
\node[anchor=east,font=\scriptsize] at (4.88,-6.70) {$-0.045$};
\node[anchor=west,font=\scriptsize] at (6.48,-6.70) {preference};

\fill[JSred!28] (6.35,-7.28) rectangle (5.33,-7.12);
\node[anchor=east,font=\scriptsize] at (5.21,-7.20) {$-0.034$};
\node[anchor=west,font=\scriptsize] at (6.48,-7.20) {relevance};

\fill[JSred] (.55,-7.85) rectangle (.95,-7.69);
\node[anchor=west,font=\scriptsize] at (1.05,-7.77) {clears the threshold};
\fill[JSred!28] (3.95,-7.85) rectangle (4.35,-7.69);
\node[anchor=west,font=\scriptsize] at (4.45,-7.77) {not established};
\draw[dashed,JSred,line width=.7pt] (6.80,-7.90) -- (6.80,-7.64);
\node[anchor=west,font=\scriptsize] at (6.95,-7.77) {declared threshold $=0.02$};

\end{tikzpicture}}
\caption{\textbf{How one item is judged, and why the endpoint is a
difference.} (a) Each item goes to the same judge four times: template A, A
again unchanged, a reworded template B, and B again.
\textcolor{JSblue}{Blue} compares a template with its own repeat, the judge's
agreement with itself and so its reproducibility ceiling;
\textcolor{JSorange}{orange} compares the two wordings. (b) Subtracting one
from the other charges decoding noise to the ceiling, so a judge that already
disagrees with itself has less left to lose and shows a \emph{smaller} gap,
which is why cells below $0.90$ are reported but not read. (c) The answer,
pooled over judges and clustered by vendor (Table~\ref{tab:vendor}). Takeaway:
rewording moves a judge's verdict by more than its own decoding noise does,
most of all when the judge is asked for a rating rather than a choice.}
\label{fig:design}
\end{figure}

Measuring this well is harder than issuing two prompts and counting
disagreements. Three challenges have to be addressed.

\textbf{i) Difficulty in separating sensitivity to wording from the judge's own
noise.} A judge disagrees with itself on the identical prompt, so raw
disagreement under rewording mixes two sources. The obvious correction,
subtracting the one from the other, cannot be trusted everywhere: a judge that
already disagrees with itself has little agreement left to lose, so its
corrected score is pulled toward zero and noise masquerades as stability. In a
pilot we discarded, a misconfigured decoding parameter produced a ceiling low
enough to absorb the entire effect we report, and nothing in the output
distinguished it from a genuine one. A usable endpoint needs the correction
\emph{and} a rule for when the corrected value may be read
(Section~\ref{sec:repeat}).

\textbf{ii) Difficulty in building a benchmark whose difficulty is not an
artifact of its own construction.} If a heuristic that ignores the construct can
pass, a high score says nothing about judgment. Always answer A. Pick the longer
candidate. Pick the passage sharing more words with the query. Yet a benchmark
nothing can pass is equally useless, and a construction that suppresses
exploitable signal can suppress genuine signal by the same mechanism. Both
directions have to be checked, and neither failure shows in labels, provenance
or aggregate statistics: every defect we found in our own rebuilds had passed
each check that was not looking for it (Section~\ref{sec:benchmark}).

\textbf{iii) Difficulty in drawing a claim from judges that are not independent
draws.} Checkpoints from one vendor share pretraining corpora, tuning recipes
and decoding defaults, so a sensitivity learned once reappears across siblings
and is counted once per sibling. Treating them as separate evidence is what
produced the claim this work has since retracted. Counting them honestly costs
precision, and the cost is large enough to change conclusions
(Section~\ref{sec:pooling}).

\paragraph{This paper is a correction as well as a contribution.}
An earlier version of this benchmark
\ifblind(reference withheld for review)\else\citep{bellibatlu2026judgesense}\fi{}
was withdrawn after review. Its 75 items were hardcoded in source while being
described as benchmark-sourced; it presented the correct candidate first in
every pairwise item, so a judge that always answered ``A'' was
indistinguishable from a perfect one; and one factuality template inverted the
answer polarity, inflating measured disagreement uniformly across every judge.
We rebuilt from real corpora with per-item provenance, and the rebuild had
further defects of the same class: a template assignment that functioned as an
answer key, candidate responses carrying contradictory gold labels, and a
length-balancing step that installed a difficulty confound in place of the
verbosity one. We report those too. None is exotic, and each survived every
check that was not specifically looking for it.

\paragraph{Contributions.}
To tackle the above challenges, this paper introduces \textsc{JudgeSense}. We
state first what is not ours: the agreement rate in Equation~\ref{eq:jss} is
what \citet{elazar2021measuring} use for consistency in pretrained models, what
\citet{arabzadeh2025human} compute against relevance labels, and what
\citet{zhuo2024prosa} express as a variance, while \citet{ahn2025prompt}
distinguish inconsistency under repetition from inconsistency under rewording.
No single prior source covers all of what follows; we borrow the measure and the
distinction, and supply the rest.

\begin{itemize}\itemsep2pt
\item \textbf{An endpoint measured against the judge's own noise (i).} We
  subtract repeat agreement from paraphrase agreement, scoring each judge
  against its own reproducibility ceiling, and decline to read any cell whose
  ceiling is too low to carry the comparison.
\item \textbf{A benchmark whose attack surface is measured, not asserted (ii).}
  880 items from four human-labelled corpora with per-item provenance, released
  with the heuristic baselines that would beat it, their intervals and support
  counts, and a declared threshold the best judge must clear for a task to count
  as discriminating.
\item \textbf{Inference that does not count siblings twice (iii).} The
  confirmatory contrast clusters judges by the vendor that trained the
  checkpoint. Two of our four claims do not survive it, and we say so.
\item \textbf{Refusal treated as a measurement.} Provider-flagged declines are
  recorded apart from unparseable output: alike once parsing has failed, but
  different facts about a judge.
\item \textbf{Evidence that the ordinal task is the least stable.} Coherence carries the largest pooled effect of the four, roughly
three times the next largest, still the largest when its scale is collapsed to
binary, and the largest loss for fourteen of the sixteen judges whose four
cells are all readable. It is also where judges differ most from each other:
the least stable task and the least agreed-upon one at once.
\end{itemize}

\section{Related work}
\label{sec:related}

\paragraph{LLM judges and their validation.}
\citet{zheng2023judging} established the practice at scale through MT-Bench and
Chatbot Arena, and holistic evaluation efforts adopted it
\citep{liang2023holistic}. Validation is almost always correlational: a judge
is trusted when its scores agree with human ratings in aggregate
\citep{chiang2023can, liu2023geval}. \citet{chiang2023can} find that agreement
broadly positive but task-dependent, which prefigures a result below. What
correlational validation cannot say is whether rerunning an evaluation under
differently worded instructions will reproduce it.

\paragraph{Known judge biases.}
\citet{wang2023fair} show position bias, the preference for whichever candidate
is presented first, to be strong enough to require explicit correction.
\citet{shi2024judging} extend that across twelve judges and more than
100{,}000 evaluation instances, and find it a stable, model-specific property
that bites hardest when the two candidates are close in quality. Length is the
other documented confound \citep{wang2023fair}. Both describe what judges
prefer, and Section~\ref{sec:benchmark} treats them as constraints on
construction. Neither reports how a verdict moves when the instruction itself
is reworded.

\paragraph{Prompt sensitivity in language models generally.}
Formatting choices alone swing accuracy by as much as 76 points on some
benchmarks \citep{sclar2024quantifying}, paraphrasing task instructions alters
model rankings \citep{mizrahi2024state}, and consistency under rephrasing has
been treated as a first-class property of pretrained models
\citep{elazar2021measuring}. The sensitivity has itself been benchmarked \citep{zhuo2024prosa,
razavi2025benchmarking, romanou2026brittlebench}:
\citet{razavi2025benchmarking} report that current methods cannot anticipate
which prompts will trip a model up, which argues for measuring it directly, and
\citet{romanou2026brittlebench} attribute up to half of a model's performance
variance to semantics-preserving perturbations. \citet{wei2024systematic} study
prompt-template effects on judge reliability, restricted to alignment tasks and
under a different metric.

Separating the two inconsistencies is not our idea. \citet{ahn2025prompt} name
them: inconsistency under repetition, a property of decoding, and inconsistency
under rewording, a property of the prompt. We subtract the first from the
second, report the difference, and treat the first as a precondition for reading
the second at all. \citet{weng2026policy} apply the same correction elsewhere,
counting verdict flips on safety rubrics only where they exceed a judge's own
baseline jitter.

All of this concerns a model being tested. A judge does not generate an answer;
it returns a label from a small finite set, so its sensitivity is a question
about decision stability, not output fluency, and these measurements do not
transfer to the model doing the testing.

\paragraph{Judging judges.}
\citet{thakur2024judging} assess thirteen judges on a clean alignment task and
document sensitivity to prompt complexity, length and other surface features
alongside a general tendency toward leniency; \citet{li2024llms} survey the area
broadly. The last two years have made judge reliability a subject in its own right.
\citet{li2025cannot} find reliability highly sensitive to the choice of judge
prompt template; \citet{norman2026reliability} evaluate twenty-one judges and
show that exact-match agreement overstates discrimination once corrected for
chance, which is why we report Cohen's $\kappa$ beside every raw figure; and
\citet{feng2025assessing} score judges on self-consistency without human
annotation, as we do. Closest in method is \citet{yagubyan2026coin}, who repeat
identical evaluations fifty times per question and report that semantically
equivalent templates change the majority outcome in a quarter of cases, on two
judges from one provider.

Closest in question is \citet{arabzadeh2025human}, who compare human- and
LLM-authored prompt variants for relevance judgment (90 prompts, three judge
models) against TREC labels using Cohen's $\kappa$, the agreement rate
discounted for what two raters would reach by chance. The difference is the
referent: that work tracks how rewording moves a judge's \emph{accuracy} against
human gold, while JSS never consults the gold label and tracks how rewording
moves a judge's \emph{agreement with itself}. The two diverge when a judge flips
verdicts in compensating directions, which our measurement sees and an
accuracy-referenced one does not.

\paragraph{Why not simply evaluate under many prompts?}
\citet{hua2025flaw} dispute the premise, arguing that much reported prompt
sensitivity is an artifact of log-likelihood scoring and rigid answer matching,
and that an LLM judge in their place cuts variance across templates. We read
that as motivating this paper: their comparison is of ways to score a model,
ours is of the scorer. If a judge absorbs the benchmark's prompt sensitivity,
how much the judge itself carries is the question that remains, and it is
answered by measuring rather than by assuming the substitution is free. Our
parsing is strict and our decision spaces small and fixed, so the leniency their
critique targets cannot inflate our estimate.

\citet{mizrahi2024state} advocate multi-prompt evaluation as a methodological
default. It is good advice for the evaluation being run, but it must be re-run
for every new judge over a prompt distribution chosen for that study, whereas a
fixed paraphrase set yields a number computed once and portable anywhere.
Variance-based characterisations \citep{sclar2024quantifying, zhuo2024prosa}
need gold labels, since they measure how accuracy moves across prompt variants;
ours needs none, so it applies where no gold labels exist, which is the
usual situation. Ensemble judging \citep{li2024llms} manages
sensitivity by combining judges, and says nothing about whether any individual
judge is stable enough to use alone.

In contrast to existing work, this paper measures the stability of the judge
itself, against its own noise rather than against a gold label, across
twenty-five judges from six providers, and reports the result under an
independence assumption weak enough that two of its four claims do not survive.

\paragraph{Sources.}
The items come from existing human-labelled corpora: TruthfulQA for factual
accuracy \citep{lin2022truthfulqa}, SummEval for expert coherence ratings
\citep{fabbri2021summeval}, BEIR's TREC-COVID split for graded relevance
judgements \citep{thakur2021beir}, and MT-Bench's human preference votes
\citep{zheng2023judging}. No label in this benchmark is assigned or inferred by
us. The judges are drawn from current instruction-tuned and reasoning-tuned
families \citep{openai2023gpt4, gemini2023gemini, grattafiori2024llama3,
qwen2024qwen25, jiang2023mistral, deepseek2025r1}.

\section{The JudgeSense benchmark}
\label{sec:benchmark}

JudgeSense measures whether an LLM judge returns the same verdict when the
same evaluation request is worded differently. A judge whose verdict depends on
which of two equivalent phrasings it received is not usable as a measuring
instrument, however accurate it looks on average.

\subsection{Tasks and sources}
\label{sec:sources}

The benchmark covers four evaluation tasks, each drawn from an existing
human-labelled corpus. No item is authored by us and no label is assigned or
inferred by us; every label is the one the source corpus already carried.

\begin{description}
  \item[Factuality] (250 items, TruthfulQA). A statement is judged accurate or
inaccurate. Items whose question is ill-posed, and whose only defensible
answer is a refusal to answer, are excluded at load time.
  \item[Coherence] (250 items, SummEval). A summary is rated 1--5 for
coherence against expert annotations. Summaries with identical text but
conflicting expert labels are dropped, not merged.
  \item[Relevance] (250 items, 500 rows, BEIR TREC-COVID). Given a topic, the
judge picks the more relevant of two passages. \emph{Both} candidates carry a
human relevance grade: the positive is graded fully relevant (2) and the
distractor is one a human assessor explicitly rejected (0). Partially relevant
documents (grade 1) are used for neither role.
  \item[Preference] (130 items, 260 rows, MT-Bench human judgments). The judge
picks the response humans preferred. Only comparisons with at least two
annotator votes and a clear majority are used.
\end{description}

The relevance construction departs from the usual practice of drawing
distractors from a positives-only collection. There, a negative can only be a
document \emph{absent} from an incomplete judgment set, so a topically similar
distractor may in fact be relevant but unjudged, and the item penalises a
defensible answer. TREC-COVID carries tens of thousands of explicit
non-relevant assessments, so both candidates here are human-labelled and the
distractor is one a human affirmatively rejected.

Every shipped item records its source dataset, split, per-record identifier and
source configuration, so any item can be traced to the row it came from. When a
source is unavailable the loader raises; there is no fallback path that could
quietly generate an item in its place.

\subsection{Paraphrase design}

Each item is issued under two of five instruction templates, intended to differ
in wording and not in what they ask; all twenty are printed in full in
Appendix~\ref{app:templates}. That intent is a construct assumption: if
it fails, a judge answering both arms correctly is scored as unstable, and the
result is semantic drift, not prompt sensitivity.

We audit the released files, not the generator, recovering each template from
the shipped prompts and the payload beside them. The audit is offline,
deterministic, runs in continuous integration, and enforces four properties.
\emph{Label space}: both templates in a pair request the same answer set
(\{YES, NO\} for factuality, $\{1,\dots,5\}$ for coherence, \{A, B\} for the
pairwise tasks). \emph{Polarity}: no answer token is bound to opposite
conclusions, and no template asks the inverted question. \emph{Construct}:
every template draws its vocabulary from its own task. \emph{Non-triviality}:
no pair is a near-duplicate. All four hold on all twenty shipped templates.

\paragraph{The limits of the audit.}
Those properties are structural. None shows that two differently worded
instructions ask for the same \emph{judgement}, the assumption the benchmark
rests on. That assumption is smaller than the item count suggests, and its size
is what makes it checkable: equivalence is a property of the instruction pair,
not of the item, and within an item both arms carry byte-identical content, so
no particular passage can make one arm ask a different question. The $880$
items draw on ten instruction pairs per task, forty in all, built from the
twenty templates printed in Appendix~\ref{app:templates}, so a reader can audit
the assumption directly against the four decision spaces.

We did not have those forty pairs scored by an independent judge, human or
model, for this release. The manual records from the first version are keyed to
identifiers no item here uses. A model-based check returned unanimous
agreement, but its validators consulted the generator and the equivalence tests
instead of judging the pairs as shown; that is circular, and we rest nothing on
it.

Two further properties we measure and report but cannot enforce. Templates
place the payload at different points: in relevance, T1 opens the query at
$46.4\%$ of the template against T3 at $9.0\%$, a within-pair divergence of
$37.4$ percentage points; and reading difficulty varies by up to $3.7$
Flesch--Kincaid grade levels within a pair. Either could make one arm
intrinsically harder. Three residuals remain open: the audit groups surface
forms into one construct by an authored mapping, so whether ``factually
correct'' and ``accurate'' are synonyms for a given judge is an assumption;
implicit thresholds may shift between phrasings even when the construct does
not; and tokenisation or pretraining frequency may favour one phrasing for one
model. Semantic drift is reduced as an alternative explanation, not
eliminated.

\paragraph{A limitation of the assignment.}
Each item is issued under exactly one of the ten unordered template pairs,
balanced across pairs, so template identity nests perfectly inside item
identity: the items behind $T_i$ and $T_j$ are disjoint, and any apparent
difference between two templates is equally a difference between the items they
were shown. Claims of the form ``template A is more stable than template B''
are not identifiable here, and we make none. The within-item contrast the
design was built for is unaffected, since the two arms of a pair differ only in
wording with the item held fixed. Appendix~\ref{app:polarity} records the
polarity defect this design corrects.

\subsection{Position control}

Both pairwise tasks present every item twice, once in each candidate ordering.
The correct answer therefore sits at position A in exactly half the rows, and
a judge that ignores content and always answers ``A'' scores $0.500$, not
$1.000$.

This corrects an earlier version in which the correct candidate always
appeared first, where most judges scored a degenerate
$\mathrm{JSS}=1.000$ that reflected the layout and no judgment at all.

\subsection{Release and integrity}

The dataset is released with Croissant metadata, per-item provenance, and a
content hash for each split that excludes retrieval timestamps, so a rebuild
can be verified as identical in content without being byte-identical in
metadata. A 32-check audit gate (eight checks over each of four splits) runs
in continuous integration and fails the build on duplicate content, unbacked
provenance claims, label degeneracy, insufficient clustering support,
contradictory ground truth, or a length shortcut exceeding its bound. A check
that cannot read the data it is meant to inspect fails; it does not pass. A
check that silently measures nothing therefore cannot be mistaken for one that
found nothing wrong.
\section{Measuring sensitivity}
\label{sec:metrics}

This section settles three things before any number is reported: what is
measured on a single cell, how the cells are pooled into a claim about a task,
and which cells are read at all. The first is the endpoint,
$\Delta\mathrm{JSS}$, defined in Section~\ref{sec:repeat} as the gap between
a judge's agreement across two wordings and its agreement with itself. The
second is a choice about independence, since sibling checkpoints from one
vendor are not separate evidence (Section~\ref{sec:pooling}). The third is a
threshold on the repeat arm, below which a cell is reported and not read
(Section~\ref{sec:bar}); that bar conditions on a term of the quantity it
filters, so the contrast is reported with and without it.

\subsection{Judge Sensitivity Score}

Equation~\ref{eq:jss} gives the endpoint's first term: the rate at which a
judge returns the same decision under two equivalent templates. It never
consults the ground-truth label, so accuracy is reported separately. On the
pairwise tasks the companion measure is \emph{position-corrected} accuracy,
whether the judge selected the side carrying the ground truth given that the
swap design places it at position A in exactly half the rows; on the pointwise
tasks it is ordinary accuracy against the source label. Raw agreement is also
inflated when a
judge's output distribution is compressed, since a judge answering ``YES''
almost always agrees with itself, so alongside every raw figure we give the
chance-corrected form (Cohen's $\kappa$, quadratic-weighted for the ordinal
coherence task).

As Section~\ref{sec:intro} states, the agreement rate itself is borrowed. What
is new here is the benchmark, the systematic application to the model doing the
judging, and the endpoint defined next.

\subsection{The repeat baseline, and why it is the endpoint}
\label{sec:repeat}

Raw JSS below $1.0$ does not by itself demonstrate sensitivity to
\emph{wording}. A judge sampling at nonzero temperature, or served through a
non-deterministic backend, disagrees with itself on the identical prompt, and
raw JSS attributes that decoding noise to paraphrasing. We separate the two by
issuing one template a second time, unchanged. The resulting
$\mathrm{JSS}_{\text{rep}}$ is the judge's reproducibility ceiling on that
task, and the primary endpoint is the difference

\[
\Delta\mathrm{JSS} = \mathrm{JSS}_{\text{para}} -
\mathrm{JSS}_{\text{rep}},
\]

which is negative when rewording costs more stability than resampling does.

A pre-registration is worth only what is true of it, and ours is partly
post-hoc. Commit \texttt{b1b507a}, before the full sweep, fixed the endpoint,
the estimator, the item-level clustering unit, and one decision rule: an effect
is present when the interval excludes zero. Those four preceded the sweep
though not the first API call, since a five-row-per-cell smoke wave had already
exercised the harness against three judges; it produced no reportable estimate
and none of its output enters the results. Commit \texttt{82ea522}, some
seventy minutes after the first full judge's results were committed, added the
support floors, the multiplicity plan and the discrimination thresholds. Those
are \emph{not} pre-registered, and two of the four discrimination thresholds
were set while the corresponding accuracies were already visible in the
repository, which disqualifies them as confirmatory tests; they are reported as
post-hoc descriptive checks. One correction is recorded in the plan itself: the
direction of subtraction in the endpoint's prose was stated backwards relative
to the estimator, which had been implemented and tested beforehand.

The multiplicity plan added at \texttt{82ea522} is the one this paper applies:
a pooled contrast per task with Holm correction, per-cell intervals demoted to
exploratory under Benjamini--Hochberg control at a $10\%$ false discovery rate,
and a reported minimum detectable effect. No judge is ranked against another on
the strength of an unadjusted cell. Because the plan was written after the
first judge's results were committed, the confirmatory and exploratory labels
are themselves declared rather than pre-registered, and we carry that
qualification wherever they are used.

\subsection{Pooling across judges}
\label{sec:pooling}

The confirmatory claims pool judges within a task, and the estimator is the
unweighted mean of the judge-level $\Delta\mathrm{JSS}$ values for that task's
readable cells. Weighting by precision would let the judges with the narrowest
intervals decide a cross-judge claim, which is not what we are asking.

The sampling unit is the judge, and judges are not independent: twenty-five
checkpoints come from eight vendors, one of which supplies seven. Sibling
checkpoints share pretraining corpora, instruction-tuning recipes and decoding
defaults, so a sensitivity learned once reappears across the siblings and would
be counted once per sibling. That is the error we attribute to the superseded
version of this work, whose retracted claim rested on three judges from one
vendor. \citet{kohli2026nine} put a number on the same effect for judge panels:
across nine frontier judges from seven families, correlated errors leave roughly
two votes' worth of independent information, so counting judges overstates the
evidence they jointly carry. We therefore report a cluster-robust standard error with the vendor as
the cluster, using the CR1 correction and Student's $t$ on $G-1$ degrees of
freedom instead of a normal approximation, which with eight clusters is
optimistic. Both the point-null test and the equivalence-style test against the
declared threshold are Holm-corrected over their own family of four, and the
unclustered standard error is reported alongside so the cost of the assumption
is visible.

\subsection{The readability bar, and what it costs}
\label{sec:bar}

We decline to read any cell whose repeat ceiling falls below $0.90$. The bar is
declared rather than pre-registered: it was fixed after the pilot described in
Section~\ref{sec:results}, in which a misconfigured decoding parameter produced
a ceiling of $0.864$ and absorbed the entire effect.

It also conditions on a term of the quantity it filters, since
$\Delta\mathrm{JSS}$ subtracts $\mathrm{JSS}_{\text{rep}}$ and the bar is a
threshold on $\mathrm{JSS}_{\text{rep}}$, so retaining only high-ceiling cells
is not a neutral quality gate. We report the contrast both ways.
Dropping the bar moves coherence from $-0.181$ to $-0.158$ and relevance from
$-0.034$ to $-0.039$; factuality is unchanged at $-0.057$ and preference moves
from $-0.045$ to $-0.042$. The bar flatters the largest effect and penalises
the smallest, the direction that favours our own headline. Every pooled
conclusion in Section~\ref{sec:results} holds under both.

We also report a minimum detectable effect per cell: the smallest
$|\Delta\mathrm{JSS}|$ that cell could have told apart from zero at its own
support, computed as $(z_{0.975} + z_{0.80})$ times the bootstrap standard
error of its estimate, the conventional two-sided calculation at $80\%$ power.
A cell whose estimate falls below its own minimum detectable effect is
uninformative rather than null, and thirty-six of the eighty-seven readable
cells are in that position.

\subsection{Uncertainty and decoding budget}

All intervals are 95\% cluster bootstrap intervals over 2{,}000 resamples with
the item as the resampling unit. The design forces that choice: a pairwise item
contributes two rows, one per candidate ordering, and its repeat arm nests
inside it, so resampling rows independently treats correlated observations as
independent and understates uncertainty. The clustering unit is a mandatory
argument in our implementation, with no default, so it cannot be omitted by
accident. An effect is reported as present for a judge--task cell when the
interval for $\Delta\mathrm{JSS}$ excludes zero, and the direction is read from
the sign.

Every judge is run with an identical output token limit of $1{,}024$ tokens. A
class-dependent limit, tight for instruction-tuned judges and generous for
reasoning-tuned ones, would confound judge class with inference configuration,
so any difference between the classes would admit an explanation having nothing
to do with the models; it would also truncate real responses, since a judge
that emits any preamble before its verdict exhausts a tight cap mid-sentence
and the fragment is retained as its answer. Because a token limit is a ceiling
and not a target, the matched budget costs nothing for judges that emit a
single-token answer, and more only for the verbose non-compliers whose calls
were previously truncated into noise.

We quote no rate for any of this. Every committed run was made under the
matched policy, so we hold no measurement of the cap-termination rate under the
class-dependent one, and quantifying the budget's effect would require a
subsample run under that configuration. We did not run it, and we make no claim
about the effect's size; the case for the matched budget rests on the confound
being structural.

\section{Results}
\label{sec:results}

We report twenty-five judges across six providers on the four tasks, giving
ninety-eight judge--task cells with an endpoint. Two cells report none: the
Sonnet and Opus preference splits stopped below the declared support floor when
the API balance was exhausted mid-run, and Section~\ref{sec:results-limits}
states what that costs. Every figure below is produced by
\texttt{scripts/regenerate\_results.py} from the committed raw outputs; none is
transcribed.

{\footnotesize
\begin{longtable}{llrrrclrr}
\caption{\textbf{Paraphrase sensitivity per judge and task.}
$\Delta\mathrm{JSS} = \mathrm{JSS}_{\text{para}} - \mathrm{JSS}_{\text{rep}}$ is
the pre-registered endpoint, negative when rewording costs more agreement than
re-issuing the identical prompt does. The bar draws that estimate from a zero
line, leftward for a cost, with the hairline at the declared $0.02$ threshold;
bands group one judge's four tasks. Intervals are 95\% item-clustered bootstrap
over 2{,}000 resamples, and cells below the support floor of 100 clusters
report no endpoint. $r$ is the refusal rate and $m$ the malformed-output rate
over the arms the judge attempted. The first two columns do not subtract to the
printed delta wherever a cell carries refusals or a swapped-ordering imbalance,
because the delta uses the canonical ordering only. Takeaway: rewording costs
agreement almost everywhere, and coherence is the largest loss for nearly every
judge and the column they disagree on most.}\label{tab:main}\\
\toprule
\rowcolor{headBand}
Judge & Task & $n$ & JSS & $\mathrm{JSS}_{\text{rep}}$ & \multicolumn{1}{c}{\footnotesize\itshape effect} & $\Delta\mathrm{JSS}$ [95\% CI] & $r$ & $m$ \\
\midrule
\endfirsthead
\multicolumn{9}{l}{\footnotesize\itshape Table~\ref{tab:main}, continued from the previous page.}\\
\toprule
\rowcolor{headBand}
Judge & Task & $n$ & JSS & $\mathrm{JSS}_{\text{rep}}$ & \multicolumn{1}{c}{\footnotesize\itshape effect} & $\Delta\mathrm{JSS}$ [95\% CI] & $r$ & $m$ \\
\midrule
\endhead
\midrule
\multicolumn{9}{r}{\footnotesize\itshape continued overleaf}\\
\endfoot
\bottomrule
\endlastfoot
claude-haiku & factuality & 250 & 0.956 & 1.000 &\dvis{-0.044} & $-0.044$ {\scriptsize[-0.072, -0.020]} & 0.000 & 0.000 \\
 & coherence & 250 & 0.792 & 0.998 &\dvis{-0.206} & $-0.206$ {\scriptsize[-0.258, -0.154]} & 0.000 & 0.000 \\
 & relevance & 500 & 0.930 & 0.990 &\dvis{-0.058} & $-0.058$ {\scriptsize[-0.086, -0.032]} & 0.000 & 0.010 \\
 & preference & 260 & 0.896 & 0.965 &\dvis{-0.081} & $-0.081$ {\scriptsize[-0.127, -0.038]} & 0.000 & 0.031 \\
\addlinespace
\rowcolor{judgeBand}claude-opus-4-7 & factuality & 250 & 0.960 & 0.990 &\dvis{-0.030} & $-0.030$ {\scriptsize[-0.056, -0.006]} & 0.000 & 0.000 \\
\rowcolor{judgeBand} & coherence & 250 & 0.736 & 0.916 &\dvis{-0.180} & $-0.180$ {\scriptsize[-0.236, -0.128]} & 0.000 & 0.000 \\
\rowcolor{judgeBand} & relevance & 500 & 0.964 & 0.998 &\dvis{-0.036} & $-0.036$ {\scriptsize[-0.063, -0.013]} & 0.051 & 0.000 \\
\rowcolor{judgeBand} & preference & 170 & 0.963 & --- & & \emph{support below floor} & 0.000 & 0.056 \\
\addlinespace
claude-sonnet & factuality & 250 & 0.952 & 0.990 &\dvis{-0.038} & $-0.038$ {\scriptsize[-0.062, -0.018]} & 0.000 & 0.008 \\
 & coherence & 250 & 0.828 & 0.990 &\dvis{-0.162} & $-0.162$ {\scriptsize[-0.210, -0.118]} & 0.000 & 0.000 \\
 & relevance & 500 & 0.584 & 0.789 &\dvis{-0.158} & $-0.158$ {\scriptsize[-0.206, -0.110]} & 0.247 & 0.224 \\
 & preference & 44 & 0.559 & --- & & \emph{support below floor} & 0.000 & 0.432 \\
\addlinespace
\rowcolor{judgeBand}deepseek-v4-flash & factuality & 250 & 0.888 & 0.908 &\dvis{-0.020} & $-0.020$ {\scriptsize[-0.056, 0.016]} & 0.000 & 0.010 \\
\rowcolor{judgeBand} & coherence & 250 & 0.616 & 0.670 &\dvis{-0.054} & $-0.054$ {\scriptsize[-0.116, 0.008]} & 0.000 & 0.000 \\
\rowcolor{judgeBand} & relevance & 500 & 0.944 & 0.962 &\dvis{-0.006} & $-0.006$ {\scriptsize[-0.026, 0.012]} & 0.000 & 0.004 \\
\rowcolor{judgeBand} & preference & 260 & 0.911 & 0.923 &\dvis{-0.023} & $-0.023$ {\scriptsize[-0.058, 0.012]} & 0.000 & 0.025 \\
\addlinespace
deepseek-v4-flash-ds & factuality & 250 & 0.888 & 0.900 &\dvis{-0.012} & $-0.012$ {\scriptsize[-0.046, 0.022]} & 0.000 & 0.004 \\
 & coherence & 250 & 0.628 & 0.630 &\dvis{-0.002} & $-0.002$ {\scriptsize[-0.062, 0.060]} & 0.000 & 0.004 \\
 & relevance & 500 & 0.946 & 0.952 &\dvis{-0.008} & $-0.008$ {\scriptsize[-0.028, 0.012]} & 0.000 & 0.016 \\
 & preference & 260 & 0.892 & 0.900 &\dvis{+0.000} & $+0.000$ {\scriptsize[-0.038, 0.038]} & 0.000 & 0.050 \\
\addlinespace
\rowcolor{judgeBand}deepseek-v4-pro & factuality & 250 & 0.904 & 0.922 &\dvis{-0.018} & $-0.018$ {\scriptsize[-0.046, 0.010]} & 0.000 & 0.042 \\
\rowcolor{judgeBand} & coherence & 250 & 0.720 & 0.742 &\dvis{-0.022} & $-0.022$ {\scriptsize[-0.076, 0.032]} & 0.000 & 0.004 \\
\rowcolor{judgeBand} & relevance & 500 & 0.960 & 0.968 &\dvis{+0.008} & $+0.008$ {\scriptsize[-0.004, 0.022]} & 0.000 & 0.014 \\
\rowcolor{judgeBand} & preference & 260 & 0.923 & 0.896 &\dvis{+0.019} & $+0.019$ {\scriptsize[-0.012, 0.058]} & 0.000 & 0.048 \\
\addlinespace
gemini-3.7-flash & factuality & 250 & 0.976 & 0.992 &\dvis{-0.016} & $-0.016$ {\scriptsize[-0.032, -0.004]} & 0.000 & 0.000 \\
 & coherence & 250 & 0.840 & 0.902 &\dvis{-0.062} & $-0.062$ {\scriptsize[-0.102, -0.022]} & 0.000 & 0.000 \\
 & relevance & 500 & 0.982 & 0.992 &\dvis{-0.012} & $-0.012$ {\scriptsize[-0.026, 0.000]} & 0.000 & 0.000 \\
 & preference & 260 & 0.969 & 0.988 &\dvis{-0.019} & $-0.019$ {\scriptsize[-0.046, 0.000]} & 0.000 & 0.000 \\
\addlinespace
\rowcolor{judgeBand}gemini-flash & factuality & 250 & 0.896 & 0.996 &\dvis{-0.100} & $-0.100$ {\scriptsize[-0.136, -0.064]} & 0.000 & 0.000 \\
\rowcolor{judgeBand} & coherence & 250 & 0.688 & 0.986 &\dvis{-0.298} & $-0.298$ {\scriptsize[-0.358, -0.242]} & 0.000 & 0.000 \\
\rowcolor{judgeBand} & relevance & 500 & 0.878 & 0.950 &\dvis{-0.058} & $-0.058$ {\scriptsize[-0.084, -0.034]} & 0.000 & 0.056 \\
\rowcolor{judgeBand} & preference & 260 & 0.935 & 0.996 &\dvis{-0.027} & $-0.027$ {\scriptsize[-0.058, -0.004]} & 0.000 & 0.002 \\
\addlinespace
gemma-4-31b & factuality & 250 & 0.980 & 0.992 &\dvis{-0.012} & $-0.012$ {\scriptsize[-0.032, 0.004]} & 0.000 & 0.000 \\
 & coherence & 250 & 0.848 & 0.958 &\dvis{-0.110} & $-0.110$ {\scriptsize[-0.156, -0.070]} & 0.000 & 0.000 \\
 & relevance & 500 & 0.866 & 0.904 &\dvis{-0.040} & $-0.040$ {\scriptsize[-0.062, -0.020]} & 0.000 & 0.092 \\
 & preference & 260 & 0.996 & 0.996 &\dvis{-0.004} & $-0.004$ {\scriptsize[-0.023, 0.012]} & 0.000 & 0.000 \\
\addlinespace
\rowcolor{judgeBand}glm-5.2 & factuality & 250 & 0.880 & 0.910 &\dvis{-0.030} & $-0.030$ {\scriptsize[-0.056, -0.004]} & 0.000 & 0.048 \\
\rowcolor{judgeBand} & coherence & 250 & 0.616 & 0.670 &\dvis{-0.054} & $-0.054$ {\scriptsize[-0.118, 0.012]} & 0.000 & 0.002 \\
\rowcolor{judgeBand} & relevance & 500 & 0.950 & 0.970 &\dvis{+0.002} & $+0.002$ {\scriptsize[-0.018, 0.022]} & 0.000 & 0.007 \\
\rowcolor{judgeBand} & preference & 260 & 0.873 & 0.881 &\dvis{-0.012} & $-0.012$ {\scriptsize[-0.042, 0.019]} & 0.000 & 0.067 \\
\addlinespace
kimi-k3 & factuality & 250 & 0.948 & 0.992 &\dvis{-0.044} & $-0.044$ {\scriptsize[-0.072, -0.018]} & 0.000 & 0.000 \\
 & coherence & 250 & 0.832 & 0.904 &\dvis{-0.072} & $-0.072$ {\scriptsize[-0.118, -0.028]} & 0.000 & 0.000 \\
 & relevance & 500 & 0.972 & 0.994 &\dvis{-0.014} & $-0.014$ {\scriptsize[-0.030, -0.002]} & 0.000 & 0.000 \\
 & preference & 260 & 0.965 & 0.992 &\dvis{-0.015} & $-0.015$ {\scriptsize[-0.042, 0.004]} & 0.000 & 0.000 \\
\addlinespace
\rowcolor{judgeBand}llama-3.3-70b & factuality & 250 & 0.952 & 1.000 &\dvis{-0.048} & $-0.048$ {\scriptsize[-0.076, -0.024]} & 0.000 & 0.000 \\
\rowcolor{judgeBand} & coherence & 250 & 0.772 & 0.992 &\dvis{-0.220} & $-0.220$ {\scriptsize[-0.274, -0.170]} & 0.000 & 0.000 \\
\rowcolor{judgeBand} & relevance & 500 & 0.890 & 0.926 &\dvis{-0.046} & $-0.046$ {\scriptsize[-0.072, -0.022]} & 0.000 & 0.058 \\
\rowcolor{judgeBand} & preference & 260 & 0.965 & 1.000 &\dvis{-0.015} & $-0.015$ {\scriptsize[-0.038, 0.000]} & 0.000 & 0.004 \\
\addlinespace
llama-4-maverick & factuality & 250 & 0.916 & 0.996 &\dvis{-0.080} & $-0.080$ {\scriptsize[-0.116, -0.048]} & 0.000 & 0.000 \\
 & coherence & 250 & 0.764 & 0.986 &\dvis{-0.222} & $-0.222$ {\scriptsize[-0.274, -0.170]} & 0.000 & 0.000 \\
 & relevance & 500 & 0.962 & 0.998 &\dvis{-0.022} & $-0.022$ {\scriptsize[-0.044, -0.006]} & 0.000 & 0.006 \\
 & preference & 260 & 0.942 & 0.988 &\dvis{-0.058} & $-0.058$ {\scriptsize[-0.104, -0.019]} & 0.000 & 0.002 \\
\addlinespace
\rowcolor{judgeBand}llama-4-scout & factuality & 250 & 0.892 & 0.994 &\dvis{-0.102} & $-0.102$ {\scriptsize[-0.142, -0.066]} & 0.000 & 0.000 \\
\rowcolor{judgeBand} & coherence & 250 & 0.644 & 0.974 &\dvis{-0.330} & $-0.330$ {\scriptsize[-0.390, -0.272]} & 0.000 & 0.000 \\
\rowcolor{judgeBand} & relevance & 500 & 0.838 & 0.974 &\dvis{-0.070} & $-0.070$ {\scriptsize[-0.104, -0.040]} & 0.000 & 0.035 \\
\rowcolor{judgeBand} & preference & 260 & 0.846 & 0.938 &\dvis{-0.131} & $-0.131$ {\scriptsize[-0.192, -0.077]} & 0.000 & 0.027 \\
\addlinespace
llama3-8b & factuality & 250 & 0.884 & 0.998 &\dvis{-0.114} & $-0.114$ {\scriptsize[-0.156, -0.074]} & 0.000 & 0.000 \\
 & coherence & 250 & 0.668 & 1.000 &\dvis{-0.332} & $-0.332$ {\scriptsize[-0.392, -0.272]} & 0.000 & 0.000 \\
 & relevance & 500 & 0.896 & 0.992 &\dvis{-0.072} & $-0.072$ {\scriptsize[-0.106, -0.044]} & 0.000 & 0.005 \\
 & preference & 260 & 0.911 & 0.981 &\dvis{-0.058} & $-0.058$ {\scriptsize[-0.100, -0.023]} & 0.000 & 0.015 \\
\addlinespace
\rowcolor{judgeBand}magistral-small & factuality & 250 & 0.804 & 0.944 &\dvis{-0.140} & $-0.140$ {\scriptsize[-0.190, -0.094]} & 0.000 & 0.000 \\
\rowcolor{judgeBand} & coherence & 250 & 0.820 & 0.944 &\dvis{-0.124} & $-0.124$ {\scriptsize[-0.170, -0.080]} & 0.000 & 0.000 \\
\rowcolor{judgeBand} & relevance & 500 & 0.880 & 0.936 &\dvis{-0.088} & $-0.088$ {\scriptsize[-0.130, -0.048]} & 0.000 & 0.002 \\
\rowcolor{judgeBand} & preference & 260 & 0.846 & 0.935 &\dvis{-0.112} & $-0.112$ {\scriptsize[-0.173, -0.054]} & 0.000 & 0.000 \\
\addlinespace
mistral-medium & factuality & 250 & 0.888 & 0.980 &\dvis{-0.092} & $-0.092$ {\scriptsize[-0.130, -0.056]} & 0.000 & 0.000 \\
 & coherence & 250 & 0.892 & 0.978 &\dvis{-0.086} & $-0.086$ {\scriptsize[-0.124, -0.052]} & 0.000 & 0.000 \\
 & relevance & 500 & 0.946 & 0.976 &\dvis{-0.060} & $-0.060$ {\scriptsize[-0.094, -0.028]} & 0.000 & 0.000 \\
 & preference & 260 & 0.927 & 0.985 &\dvis{-0.069} & $-0.069$ {\scriptsize[-0.112, -0.031]} & 0.000 & 0.004 \\
\addlinespace
\rowcolor{judgeBand}mistral-small & factuality & 250 & 0.816 & 0.930 &\dvis{-0.114} & $-0.114$ {\scriptsize[-0.162, -0.066]} & 0.000 & 0.000 \\
\rowcolor{judgeBand} & coherence & 250 & 0.804 & 0.930 &\dvis{-0.126} & $-0.126$ {\scriptsize[-0.172, -0.080]} & 0.000 & 0.000 \\
\rowcolor{judgeBand} & relevance & 500 & 0.840 & 0.922 &\dvis{-0.114} & $-0.114$ {\scriptsize[-0.164, -0.068]} & 0.000 & 0.001 \\
\rowcolor{judgeBand} & preference & 260 & 0.839 & 0.923 &\dvis{-0.100} & $-0.100$ {\scriptsize[-0.173, -0.035]} & 0.000 & 0.000 \\
\addlinespace
qwen & factuality & 250 & 0.908 & 1.000 &\dvis{-0.092} & $-0.092$ {\scriptsize[-0.128, -0.056]} & 0.000 & 0.000 \\
 & coherence & 250 & 0.860 & 0.988 &\dvis{-0.128} & $-0.128$ {\scriptsize[-0.170, -0.086]} & 0.000 & 0.000 \\
 & relevance & 500 & 0.940 & 0.974 &\dvis{-0.026} & $-0.026$ {\scriptsize[-0.046, -0.008]} & 0.000 & 0.029 \\
 & preference & 260 & 0.965 & 1.000 &\dvis{-0.031} & $-0.031$ {\scriptsize[-0.062, -0.008]} & 0.000 & 0.000 \\
\addlinespace
\rowcolor{judgeBand}qwen-3.6-flash & factuality & 250 & 0.932 & 0.970 &\dvis{-0.038} & $-0.038$ {\scriptsize[-0.068, -0.012]} & 0.000 & 0.000 \\
\rowcolor{judgeBand} & coherence & 250 & 0.760 & 0.912 &\dvis{-0.152} & $-0.152$ {\scriptsize[-0.202, -0.104]} & 0.000 & 0.000 \\
\rowcolor{judgeBand} & relevance & 500 & 0.944 & 0.986 &\dvis{+0.002} & $+0.002$ {\scriptsize[-0.014, 0.018]} & 0.000 & 0.004 \\
\rowcolor{judgeBand} & preference & 260 & 0.973 & 0.992 &\dvis{-0.031} & $-0.031$ {\scriptsize[-0.065, -0.004]} & 0.000 & 0.002 \\
\addlinespace
qwen3-14b & factuality & 250 & 0.888 & 0.930 &\dvis{-0.042} & $-0.042$ {\scriptsize[-0.076, -0.010]} & 0.000 & 0.006 \\
 & coherence & 250 & 0.576 & 0.780 &\dvis{-0.204} & $-0.204$ {\scriptsize[-0.264, -0.144]} & 0.000 & 0.000 \\
 & relevance & 500 & 0.932 & 0.960 &\dvis{-0.004} & $-0.004$ {\scriptsize[-0.020, 0.012]} & 0.000 & 0.028 \\
 & preference & 260 & 0.869 & 0.912 &\dvis{-0.050} & $-0.050$ {\scriptsize[-0.085, -0.019]} & 0.000 & 0.064 \\
\addlinespace
\rowcolor{judgeBand}qwen3-32b & factuality & 250 & 0.952 & 0.996 &\dvis{-0.044} & $-0.044$ {\scriptsize[-0.072, -0.020]} & 0.000 & 0.000 \\
\rowcolor{judgeBand} & coherence & 250 & 0.704 & 0.976 &\dvis{-0.272} & $-0.272$ {\scriptsize[-0.328, -0.216]} & 0.000 & 0.000 \\
\rowcolor{judgeBand} & relevance & 500 & 0.952 & 0.992 &\dvis{-0.036} & $-0.036$ {\scriptsize[-0.062, -0.014]} & 0.000 & 0.007 \\
\rowcolor{judgeBand} & preference & 260 & 0.954 & 0.996 &\dvis{-0.027} & $-0.027$ {\scriptsize[-0.062, 0.000]} & 0.000 & 0.000 \\
\addlinespace
qwen3-8b & factuality & 250 & 0.852 & 0.912 &\dvis{-0.060} & $-0.060$ {\scriptsize[-0.098, -0.020]} & 0.000 & 0.008 \\
 & coherence & 250 & 0.560 & 0.728 &\dvis{-0.168} & $-0.168$ {\scriptsize[-0.232, -0.106]} & 0.000 & 0.008 \\
 & relevance & 500 & 0.870 & 0.908 &\dvis{-0.032} & $-0.032$ {\scriptsize[-0.050, -0.014]} & 0.000 & 0.082 \\
 & preference & 260 & 0.769 & 0.827 &\dvis{-0.065} & $-0.065$ {\scriptsize[-0.115, -0.023]} & 0.000 & 0.112 \\
\addlinespace
\rowcolor{judgeBand}qwen3.7-flash & factuality & 250 & 0.960 & 0.990 &\dvis{-0.030} & $-0.030$ {\scriptsize[-0.054, -0.008]} & 0.000 & 0.000 \\
\rowcolor{judgeBand} & coherence & 250 & 0.788 & 0.936 &\dvis{-0.148} & $-0.148$ {\scriptsize[-0.196, -0.102]} & 0.000 & 0.000 \\
\rowcolor{judgeBand} & relevance & 500 & 0.954 & 0.986 &\dvis{-0.010} & $-0.010$ {\scriptsize[-0.026, 0.004]} & 0.000 & 0.001 \\
\rowcolor{judgeBand} & preference & 260 & 0.985 & 1.000 &\dvis{-0.015} & $-0.015$ {\scriptsize[-0.038, 0.000]} & 0.000 & 0.000 \\
\addlinespace
qwen3.8-27b-hf & factuality & 250 & 0.792 & 0.852 &\dvis{-0.060} & $-0.060$ {\scriptsize[-0.090, -0.030]} & 0.000 & 0.114 \\
 & coherence & 250 & 0.760 & 0.968 &\dvis{-0.208} & $-0.208$ {\scriptsize[-0.260, -0.156]} & 0.000 & 0.002 \\
 & relevance & 500 & 0.946 & 0.980 &\dvis{-0.024} & $-0.024$ {\scriptsize[-0.046, -0.004]} & 0.000 & 0.011 \\
 & preference & 260 & 0.911 & 0.938 &\dvis{-0.031} & $-0.031$ {\scriptsize[-0.065, 0.000]} & 0.000 & 0.038 \\
\end{longtable}
}

\subsection{Headline findings}
\label{sec:findings}

Six findings, ordered by how much each bears on choosing a judge. The analyses
behind them follow in Sections~\ref{sec:results-sesoi} onward.

\paragraph{Finding 1: Rewording costs agreement on all four tasks, and on two of them the cost clears the threshold once judges are clustered by vendor.}
The pooled $\Delta\mathrm{JSS}$ is negative on every task, from coherence
($-0.181$) down to relevance ($-0.034$). Clustering by vendor and correcting
over four tests, coherence and factuality clear the $0.02$ threshold;
preference and relevance do not (Table~\ref{tab:vendor}). Unclustered standard
errors are $1.37$ to $1.63$ times smaller and would have cleared all four, but
seven of the twenty-five judges come from one vendor, and sibling checkpoints
are not independent draws. Per cell, sixty-eight of the eighty-seven readable
cells are significant under Benjamini--Hochberg at a $10\%$ false discovery
rate, which is the exploratory reading and carries no confirmatory claim
(Section~\ref{sec:repeat}).

\paragraph{Finding 2: Coherence carries the largest effect and is the task fewest judges are accurate on.}
It has both the largest pooled loss, three times factuality, and the weakest
discrimination: six of twenty-five judges clear its declared accuracy threshold
of $0.40$, against twenty-five of twenty-five on relevance at $0.70$
(Tables~\ref{tab:vendor} and~\ref{tab:discriminate}). Quadratic-weighted
$\kappa$, which credits a near miss, spans $0.579$ to $0.941$ over the sixteen
fully readable judges, so they may agree on the ordering while exact-match
accuracy punishes placement on a five-point scale. Either way, a single
coherence score is unreliable.

\paragraph{Finding 3: Most of coherence's excess is the width of its five-point scale, not the task.}
Collapsing coherence to a binary label at four or above removes $58.6\%$ of the
effect, averaged over judges (Table~\ref{tab:binarised}). What remains still
leads the four tasks, at about $1.3$ times factuality, not three. A five-point scale offers more adjacent categories for a paraphrase to shift
a verdict into. The control is not decisive, since a
different cut point would remove a different share.

\paragraph{Finding 4: A low repeat ceiling hides wording sensitivity rather than exposing it.}
Repeat agreement ranges from $0.630$ to $1.000$ across the ninety-eight cells,
and thirty-three sit at $0.99$ or above (Figure~\ref{fig:ceiling}). Because
$\Delta\mathrm{JSS}$ subtracts repeat agreement from paraphrase agreement,
decoding noise is charged to the ceiling: a judge that disagrees with itself on
identical input has less room left to lose on reworded input, so its estimate
is pulled toward zero. The coherence extremes fit that account,
\texttt{llama3-8b} losing $0.332$ from a ceiling of $1.000$ against
\texttt{gemini-3.7-flash}'s $0.062$ from $0.902$. Eleven cells whose ceiling
falls below $0.90$, six of them coherence, are therefore reported but not read,
leaving eighty-seven. Read the ceiling column first, and treat a near-zero
estimate above a low ceiling as uninformative rather than as stability.

\paragraph{Finding 5: Parameter count does not predict stability, and recency is not the story the earlier version told.}
Within the \texttt{qwen-3} ladder, size does not order $\Delta\mathrm{JSS}$ on
any task, and on relevance the middle model is the most stable
(Table~\ref{tab:ladder}). Across families the coherence extremes are what a
capability story would predict, but fourteen of the sixteen fully readable
judges lose most on coherence whatever their size. The ladder is incomplete below 8B. Recency is not
controlled at all, and what the roster shows descriptively runs against the
earlier version's claim that the newest models are the least consistent: ranking
the sixteen fully readable judges by mean endpoint, the newest Gemini is the
most stable of them and the two newest Qwen checkpoints rank fourth and fifth,
while the older \texttt{gemini-flash} and \texttt{llama3-8b} sit near the
bottom. \texttt{llama-4-scout} is the counterexample, recent and least stable of
the sixteen. We report the pattern and claim nothing from it, because
generation is confounded with family, size and training recipe across
vendors.

\paragraph{Finding 6: The same judge measured inside an agent harness yields a smaller estimate.}
Re-measuring Haiku 4.5 through an agent harness instead of a direct API call
contracts $\Delta\mathrm{JSS}$ by $40$ to $86\%$ (Table~\ref{tab:transport}).
The judge has not become more stable. Its ceiling falls
faster under the harness than paraphrase agreement does, and the subtraction
in Finding~4 turns that fall into a smaller endpoint. Harness ceilings are far higher on factuality
than on coherence (Table~\ref{tab:harnessgrid}), so the collapse is
task-dependent, not model-dependent. One judge and one harness are a narrow
base, but what follows for practice is that an estimate of wording sensitivity
does not travel across measurement surfaces: a number obtained from the API
should be re-measured inside the harness the judge will actually run in.


\subsection{Paraphrasing costs agreement beyond decoding noise}

Table~\ref{tab:main} reports the endpoint for all ninety-eight cells that
carry one (Figure~\ref{fig:delta}); the eleven below the ceiling bar are
flagged rather than dropped.

\begin{figure}[t]
\centering
\begin{tikzpicture}
\begin{axis}[
  width=0.865\linewidth, height=5.2cm,
  xlabel={$\Delta\mathrm{JSS}$\ \ (negative: rewording costs agreement)},
  xmin=-0.37, xmax=0.055,
  xtick={-0.35,-0.30,-0.25,-0.20,-0.15,-0.10,-0.05,0},
  scaled x ticks=false,
  x tick label style={/pgf/number format/fixed, /pgf/number format/precision=2},
  ytick={4,3,2,1}, yticklabels={coherence,factuality,preference,relevance},
  ymin=0.45, ymax=4.55,
  xmajorgrids, grid style={gridGrey, line width=0.4pt},
  axis line style={ruleGrey}, tick style={ruleGrey},
  tick label style={font=\footnotesize},
  label style={font=\footnotesize},
  legend style={font=\scriptsize, draw=none, fill=none,
                at={(0.5,-0.30)}, anchor=north, legend columns=5,
                cells={anchor=west}, column sep=7pt, inner sep=1pt},
  legend image post style={scale=0.85},
  clip=false, scale only axis,
]
\draw[ruleGrey, line width=0.5pt] ({axis cs:0,0}|-{rel axis cs:0,0})
                               -- ({axis cs:0,0}|-{rel axis cs:0,1});
\draw[dashed, ruleGrey] ({axis cs:-0.02,0}|-{rel axis cs:0,0})
                     -- ({axis cs:-0.02,0}|-{rel axis cs:0,1});
\addplot[only marks, mark=*, mark size=1.7pt, draw=vAnthropic, fill=vAnthropic, forget plot] coordinates {(-0.2060,4)};
\addplot[only marks, mark=*, mark size=1.7pt, draw=vAnthropic, fill=vAnthropic, forget plot] coordinates {(-0.1800,4)};
\addplot[only marks, mark=*, mark size=1.7pt, draw=vAnthropic, fill=vAnthropic, forget plot] coordinates {(-0.1620,4)};
\addplot[only marks, mark=*, mark size=1.7pt, draw=vDeepSeek, fill=white, line width=0.7pt, forget plot] coordinates {(-0.0020,4)};
\addplot[only marks, mark=*, mark size=1.7pt, draw=vDeepSeek, fill=white, line width=0.7pt, forget plot] coordinates {(-0.0540,4)};
\addplot[only marks, mark=*, mark size=1.7pt, draw=vDeepSeek, fill=white, line width=0.7pt, forget plot] coordinates {(-0.0220,4)};
\addplot[only marks, mark=*, mark size=1.7pt, draw=vGoogle, fill=vGoogle, forget plot] coordinates {(-0.0620,4)};
\addplot[only marks, mark=*, mark size=1.7pt, draw=vGoogle, fill=vGoogle, forget plot] coordinates {(-0.2980,4)};
\addplot[only marks, mark=*, mark size=1.7pt, draw=vGoogle, fill=vGoogle, forget plot] coordinates {(-0.1100,4)};
\addplot[only marks, mark=*, mark size=1.7pt, draw=vZhipu, fill=white, line width=0.7pt, forget plot] coordinates {(-0.0540,4)};
\addplot[only marks, mark=*, mark size=1.7pt, draw=vMoonshot, fill=vMoonshot, forget plot] coordinates {(-0.0720,4)};
\addplot[only marks, mark=*, mark size=1.7pt, draw=vMeta, fill=vMeta, forget plot] coordinates {(-0.2200,4)};
\addplot[only marks, mark=*, mark size=1.7pt, draw=vMeta, fill=vMeta, forget plot] coordinates {(-0.2220,4)};
\addplot[only marks, mark=*, mark size=1.7pt, draw=vMeta, fill=vMeta, forget plot] coordinates {(-0.3300,4)};
\addplot[only marks, mark=*, mark size=1.7pt, draw=vMeta, fill=vMeta, forget plot] coordinates {(-0.3320,4)};
\addplot[only marks, mark=*, mark size=1.7pt, draw=vMistral, fill=vMistral, forget plot] coordinates {(-0.1240,4)};
\addplot[only marks, mark=*, mark size=1.7pt, draw=vMistral, fill=vMistral, forget plot] coordinates {(-0.0860,4)};
\addplot[only marks, mark=*, mark size=1.7pt, draw=vMistral, fill=vMistral, forget plot] coordinates {(-0.1260,4)};
\addplot[only marks, mark=*, mark size=1.7pt, draw=vAlibaba, fill=vAlibaba, forget plot] coordinates {(-0.1520,4)};
\addplot[only marks, mark=*, mark size=1.7pt, draw=vAlibaba, fill=white, line width=0.7pt, forget plot] coordinates {(-0.2040,4)};
\addplot[only marks, mark=*, mark size=1.7pt, draw=vAlibaba, fill=vAlibaba, forget plot] coordinates {(-0.2720,4)};
\addplot[only marks, mark=*, mark size=1.7pt, draw=vAlibaba, fill=white, line width=0.7pt, forget plot] coordinates {(-0.1680,4)};
\addplot[only marks, mark=*, mark size=1.7pt, draw=vAlibaba, fill=vAlibaba, forget plot] coordinates {(-0.1480,4)};
\addplot[only marks, mark=*, mark size=1.7pt, draw=vAlibaba, fill=vAlibaba, forget plot] coordinates {(-0.2080,4)};
\addplot[only marks, mark=*, mark size=1.7pt, draw=vAlibaba, fill=vAlibaba, forget plot] coordinates {(-0.1280,4)};
\addplot[only marks, mark=diamond*, mark size=3.6pt, draw=black, fill=black, forget plot] coordinates {(-0.1809,4)};
\addplot[only marks, mark=*, mark size=1.7pt, draw=vAnthropic, fill=vAnthropic, forget plot] coordinates {(-0.0440,3)};
\addplot[only marks, mark=*, mark size=1.7pt, draw=vAnthropic, fill=vAnthropic, forget plot] coordinates {(-0.0300,3)};
\addplot[only marks, mark=*, mark size=1.7pt, draw=vAnthropic, fill=vAnthropic, forget plot] coordinates {(-0.0380,3)};
\addplot[only marks, mark=*, mark size=1.7pt, draw=vDeepSeek, fill=vDeepSeek, forget plot] coordinates {(-0.0120,3)};
\addplot[only marks, mark=*, mark size=1.7pt, draw=vDeepSeek, fill=vDeepSeek, forget plot] coordinates {(-0.0200,3)};
\addplot[only marks, mark=*, mark size=1.7pt, draw=vDeepSeek, fill=vDeepSeek, forget plot] coordinates {(-0.0180,3)};
\addplot[only marks, mark=*, mark size=1.7pt, draw=vGoogle, fill=vGoogle, forget plot] coordinates {(-0.0160,3)};
\addplot[only marks, mark=*, mark size=1.7pt, draw=vGoogle, fill=vGoogle, forget plot] coordinates {(-0.1000,3)};
\addplot[only marks, mark=*, mark size=1.7pt, draw=vGoogle, fill=vGoogle, forget plot] coordinates {(-0.0120,3)};
\addplot[only marks, mark=*, mark size=1.7pt, draw=vZhipu, fill=vZhipu, forget plot] coordinates {(-0.0300,3)};
\addplot[only marks, mark=*, mark size=1.7pt, draw=vMoonshot, fill=vMoonshot, forget plot] coordinates {(-0.0440,3)};
\addplot[only marks, mark=*, mark size=1.7pt, draw=vMeta, fill=vMeta, forget plot] coordinates {(-0.0480,3)};
\addplot[only marks, mark=*, mark size=1.7pt, draw=vMeta, fill=vMeta, forget plot] coordinates {(-0.0800,3)};
\addplot[only marks, mark=*, mark size=1.7pt, draw=vMeta, fill=vMeta, forget plot] coordinates {(-0.1020,3)};
\addplot[only marks, mark=*, mark size=1.7pt, draw=vMeta, fill=vMeta, forget plot] coordinates {(-0.1140,3)};
\addplot[only marks, mark=*, mark size=1.7pt, draw=vMistral, fill=vMistral, forget plot] coordinates {(-0.1400,3)};
\addplot[only marks, mark=*, mark size=1.7pt, draw=vMistral, fill=vMistral, forget plot] coordinates {(-0.0920,3)};
\addplot[only marks, mark=*, mark size=1.7pt, draw=vMistral, fill=vMistral, forget plot] coordinates {(-0.1140,3)};
\addplot[only marks, mark=*, mark size=1.7pt, draw=vAlibaba, fill=vAlibaba, forget plot] coordinates {(-0.0380,3)};
\addplot[only marks, mark=*, mark size=1.7pt, draw=vAlibaba, fill=vAlibaba, forget plot] coordinates {(-0.0420,3)};
\addplot[only marks, mark=*, mark size=1.7pt, draw=vAlibaba, fill=vAlibaba, forget plot] coordinates {(-0.0440,3)};
\addplot[only marks, mark=*, mark size=1.7pt, draw=vAlibaba, fill=vAlibaba, forget plot] coordinates {(-0.0600,3)};
\addplot[only marks, mark=*, mark size=1.7pt, draw=vAlibaba, fill=vAlibaba, forget plot] coordinates {(-0.0300,3)};
\addplot[only marks, mark=*, mark size=1.7pt, draw=vAlibaba, fill=white, line width=0.7pt, forget plot] coordinates {(-0.0600,3)};
\addplot[only marks, mark=*, mark size=1.7pt, draw=vAlibaba, fill=vAlibaba, forget plot] coordinates {(-0.0920,3)};
\addplot[only marks, mark=diamond*, mark size=3.6pt, draw=black, fill=black, forget plot] coordinates {(-0.0567,3)};
\addplot[only marks, mark=*, mark size=1.7pt, draw=vAnthropic, fill=vAnthropic, forget plot] coordinates {(-0.0808,2)};
\addplot[only marks, mark=*, mark size=1.7pt, draw=vDeepSeek, fill=vDeepSeek, forget plot] coordinates {(0.0000,2)};
\addplot[only marks, mark=*, mark size=1.7pt, draw=vDeepSeek, fill=vDeepSeek, forget plot] coordinates {(-0.0231,2)};
\addplot[only marks, mark=*, mark size=1.7pt, draw=vDeepSeek, fill=white, line width=0.7pt, forget plot] coordinates {(0.0192,2)};
\addplot[only marks, mark=*, mark size=1.7pt, draw=vGoogle, fill=vGoogle, forget plot] coordinates {(-0.0192,2)};
\addplot[only marks, mark=*, mark size=1.7pt, draw=vGoogle, fill=vGoogle, forget plot] coordinates {(-0.0269,2)};
\addplot[only marks, mark=*, mark size=1.7pt, draw=vGoogle, fill=vGoogle, forget plot] coordinates {(-0.0038,2)};
\addplot[only marks, mark=*, mark size=1.7pt, draw=vZhipu, fill=white, line width=0.7pt, forget plot] coordinates {(-0.0115,2)};
\addplot[only marks, mark=*, mark size=1.7pt, draw=vMoonshot, fill=vMoonshot, forget plot] coordinates {(-0.0154,2)};
\addplot[only marks, mark=*, mark size=1.7pt, draw=vMeta, fill=vMeta, forget plot] coordinates {(-0.0154,2)};
\addplot[only marks, mark=*, mark size=1.7pt, draw=vMeta, fill=vMeta, forget plot] coordinates {(-0.0577,2)};
\addplot[only marks, mark=*, mark size=1.7pt, draw=vMeta, fill=vMeta, forget plot] coordinates {(-0.1308,2)};
\addplot[only marks, mark=*, mark size=1.7pt, draw=vMeta, fill=vMeta, forget plot] coordinates {(-0.0577,2)};
\addplot[only marks, mark=*, mark size=1.7pt, draw=vMistral, fill=vMistral, forget plot] coordinates {(-0.1115,2)};
\addplot[only marks, mark=*, mark size=1.7pt, draw=vMistral, fill=vMistral, forget plot] coordinates {(-0.0692,2)};
\addplot[only marks, mark=*, mark size=1.7pt, draw=vMistral, fill=vMistral, forget plot] coordinates {(-0.1000,2)};
\addplot[only marks, mark=*, mark size=1.7pt, draw=vAlibaba, fill=vAlibaba, forget plot] coordinates {(-0.0308,2)};
\addplot[only marks, mark=*, mark size=1.7pt, draw=vAlibaba, fill=vAlibaba, forget plot] coordinates {(-0.0500,2)};
\addplot[only marks, mark=*, mark size=1.7pt, draw=vAlibaba, fill=vAlibaba, forget plot] coordinates {(-0.0269,2)};
\addplot[only marks, mark=*, mark size=1.7pt, draw=vAlibaba, fill=white, line width=0.7pt, forget plot] coordinates {(-0.0654,2)};
\addplot[only marks, mark=*, mark size=1.7pt, draw=vAlibaba, fill=vAlibaba, forget plot] coordinates {(-0.0154,2)};
\addplot[only marks, mark=*, mark size=1.7pt, draw=vAlibaba, fill=vAlibaba, forget plot] coordinates {(-0.0308,2)};
\addplot[only marks, mark=*, mark size=1.7pt, draw=vAlibaba, fill=vAlibaba, forget plot] coordinates {(-0.0308,2)};
\addplot[only marks, mark=diamond*, mark size=3.6pt, draw=black, fill=black, forget plot] coordinates {(-0.0448,2)};
\addplot[only marks, mark=*, mark size=1.7pt, draw=vAnthropic, fill=vAnthropic, forget plot] coordinates {(-0.0580,1)};
\addplot[only marks, mark=*, mark size=1.7pt, draw=vAnthropic, fill=vAnthropic, forget plot] coordinates {(-0.0357,1)};
\addplot[only marks, mark=*, mark size=1.7pt, draw=vAnthropic, fill=white, line width=0.7pt, forget plot] coordinates {(-0.1578,1)};
\addplot[only marks, mark=*, mark size=1.7pt, draw=vDeepSeek, fill=vDeepSeek, forget plot] coordinates {(-0.0080,1)};
\addplot[only marks, mark=*, mark size=1.7pt, draw=vDeepSeek, fill=vDeepSeek, forget plot] coordinates {(-0.0060,1)};
\addplot[only marks, mark=*, mark size=1.7pt, draw=vDeepSeek, fill=vDeepSeek, forget plot] coordinates {(0.0080,1)};
\addplot[only marks, mark=*, mark size=1.7pt, draw=vGoogle, fill=vGoogle, forget plot] coordinates {(-0.0120,1)};
\addplot[only marks, mark=*, mark size=1.7pt, draw=vGoogle, fill=vGoogle, forget plot] coordinates {(-0.0580,1)};
\addplot[only marks, mark=*, mark size=1.7pt, draw=vGoogle, fill=vGoogle, forget plot] coordinates {(-0.0400,1)};
\addplot[only marks, mark=*, mark size=1.7pt, draw=vZhipu, fill=vZhipu, forget plot] coordinates {(0.0020,1)};
\addplot[only marks, mark=*, mark size=1.7pt, draw=vMoonshot, fill=vMoonshot, forget plot] coordinates {(-0.0140,1)};
\addplot[only marks, mark=*, mark size=1.7pt, draw=vMeta, fill=vMeta, forget plot] coordinates {(-0.0460,1)};
\addplot[only marks, mark=*, mark size=1.7pt, draw=vMeta, fill=vMeta, forget plot] coordinates {(-0.0220,1)};
\addplot[only marks, mark=*, mark size=1.7pt, draw=vMeta, fill=vMeta, forget plot] coordinates {(-0.0700,1)};
\addplot[only marks, mark=*, mark size=1.7pt, draw=vMeta, fill=vMeta, forget plot] coordinates {(-0.0720,1)};
\addplot[only marks, mark=*, mark size=1.7pt, draw=vMistral, fill=vMistral, forget plot] coordinates {(-0.0880,1)};
\addplot[only marks, mark=*, mark size=1.7pt, draw=vMistral, fill=vMistral, forget plot] coordinates {(-0.0600,1)};
\addplot[only marks, mark=*, mark size=1.7pt, draw=vMistral, fill=vMistral, forget plot] coordinates {(-0.1140,1)};
\addplot[only marks, mark=*, mark size=1.7pt, draw=vAlibaba, fill=vAlibaba, forget plot] coordinates {(0.0020,1)};
\addplot[only marks, mark=*, mark size=1.7pt, draw=vAlibaba, fill=vAlibaba, forget plot] coordinates {(-0.0040,1)};
\addplot[only marks, mark=*, mark size=1.7pt, draw=vAlibaba, fill=vAlibaba, forget plot] coordinates {(-0.0360,1)};
\addplot[only marks, mark=*, mark size=1.7pt, draw=vAlibaba, fill=vAlibaba, forget plot] coordinates {(-0.0320,1)};
\addplot[only marks, mark=*, mark size=1.7pt, draw=vAlibaba, fill=vAlibaba, forget plot] coordinates {(-0.0100,1)};
\addplot[only marks, mark=*, mark size=1.7pt, draw=vAlibaba, fill=vAlibaba, forget plot] coordinates {(-0.0240,1)};
\addplot[only marks, mark=*, mark size=1.7pt, draw=vAlibaba, fill=vAlibaba, forget plot] coordinates {(-0.0260,1)};
\addplot[only marks, mark=diamond*, mark size=3.6pt, draw=black, fill=black, forget plot] coordinates {(-0.0343,1)};
\addlegendimage{only marks, mark=*, mark size=1.7pt, draw=vAlibaba, fill=vAlibaba}
\addlegendentry{Alibaba}
\addlegendimage{only marks, mark=*, mark size=1.7pt, draw=vAnthropic, fill=vAnthropic}
\addlegendentry{Anthropic}
\addlegendimage{only marks, mark=*, mark size=1.7pt, draw=vDeepSeek, fill=vDeepSeek}
\addlegendentry{DeepSeek}
\addlegendimage{only marks, mark=*, mark size=1.7pt, draw=vGoogle, fill=vGoogle}
\addlegendentry{Google}
\addlegendimage{only marks, mark=*, mark size=1.7pt, draw=vMeta, fill=vMeta}
\addlegendentry{Meta}
\addlegendimage{only marks, mark=*, mark size=1.7pt, draw=vMistral, fill=vMistral}
\addlegendentry{Mistral}
\addlegendimage{only marks, mark=*, mark size=1.7pt, draw=vMoonshot, fill=vMoonshot}
\addlegendentry{Moonshot}
\addlegendimage{only marks, mark=*, mark size=1.7pt, draw=vZhipu, fill=vZhipu}
\addlegendentry{Zhipu}
\addlegendimage{only marks, mark=*, mark size=1.7pt, draw=black, fill=white, line width=0.7pt}
\addlegendentry{ceiling $<0.90$}
\addlegendimage{only marks, mark=diamond*, mark size=3.6pt, draw=black, fill=black}
\addlegendentry{pooled mean}
\end{axis}
\end{tikzpicture}
\caption{\textbf{Paraphrase cost by task across twenty-five judges.} One
marker per judge--task cell, coloured by the vendor that trained the
checkpoint, which is the unit the confirmatory contrast clusters on. Hollow
markers are the eleven cells whose ceiling falls below $0.90$. Black diamonds
are the pooled mean per task; the solid rule is zero and the dashed rule the
$-0.02$ threshold. Takeaway: the cost is negative almost everywhere, coherence
has both the largest loss and much the widest spread, and the three Anthropic
judges of the superseded version sit close together, which is what produced its
retracted claim that the task matters more than the judge.}
\label{fig:delta}
\end{figure}

The four confirmatory contrasts pool judges within a task and control the
family-wise error rate by Holm--Bonferroni at $\alpha = 0.05$
(Table~\ref{tab:pooled}); the individual cells are exploratory under
Benjamini--Hochberg, and no confirmatory claim rests on a single cell.

\begin{table}[!ht]
\centering
\small
\caption{\textbf{Confirmatory contrasts, judges pooled within a task.} Each row pools the readable cells of one task; $p$ tests a null of no paraphrase cost and is Holm--Bonferroni corrected over the four contrasts. Takeaway: rewording costs agreement on every task, not only on the ordinal one.}
\label{tab:pooled}
\begin{tabular}{lrcrrl}
\toprule
\rowcolor{headBand}
Task & Judges & \multicolumn{1}{c}{\footnotesize\itshape effect} & Mean $\Delta\mathrm{JSS}$ & $p$ & Holm \\
\midrule
coherence  & $19$ & \dvis{-0.181} & $-0.181$ & $1.4\times10^{-21}$ & reject \\
factuality & $24$ & \dvis{-0.057} & $-0.057$ & $1.6\times10^{-13}$ & reject \\
preference & $20$ & \dvis{-0.045} & $-0.045$ & $5.3\times10^{-8}$  & reject \\
relevance  & $24$ & \dvis{-0.034} & $-0.034$ & $7.9\times10^{-8}$  & reject \\
\bottomrule
\end{tabular}
\end{table}

Four rejections establish less than the tally suggests. Under the strict
policy, $\Delta\mathrm{JSS}$ estimates
$-\tfrac{1}{2}\,\mathbb{E}_{\text{item}}\big[\sum_k (p_k - q_k)^2\big]$, where
$p$ and $q$ are the judge's response distributions under the two templates.
That quantity is non-positive whenever the two prompts induce different
distributions, so a point null of exactly zero is false before any data are
collected. Rejecting it is a statement about power, and the sign test
establishes direction only. The magnitude carries the argument, and we assess
it against the smallest effect declared of interest
(Section~\ref{sec:results-sesoi}).

\begin{figure}[t]
\centering
\begin{tikzpicture}
\begin{axis}[
  width=0.875\linewidth, height=5.6cm,
  xlabel={repeat ceiling $\mathrm{JSS}_{\text{rep}}$\ \ (the judge's agreement with itself)},
  ylabel={$\Delta\mathrm{JSS}$},
  xmin=0.58, xmax=1.025, ymin=-0.37, ymax=0.055,
  xtick={0.6,0.7,0.8,0.9,1.0},
  ytick={-0.35,-0.30,-0.25,-0.20,-0.15,-0.10,-0.05,0},
  scaled ticks=false,
  tick label style={/pgf/number format/fixed, /pgf/number format/precision=2,
                    font=\footnotesize},
  label style={font=\footnotesize},
  xmajorgrids, ymajorgrids, grid style={gridGrey, line width=0.4pt},
  axis line style={ruleGrey}, tick style={ruleGrey},
  legend style={font=\scriptsize, draw=none, fill=none,
                at={(0.5,-0.30)}, anchor=north, legend columns=5,
                cells={anchor=west}, column sep=7pt, inner sep=1pt},
  legend image post style={scale=0.85},
  clip=false, scale only axis,
]
\draw[dashed, ruleGrey, line width=0.6pt] (axis cs:0.90,-0.37) -- (axis cs:0.90,0.055);
\node[font=\scriptsize, text=ruleGrey, anchor=south east, inner sep=1.5pt]
  at (axis cs:0.895,-0.37) {not read};
\draw[dashed, ruleGrey] (axis cs:0.58,-0.02) -- (axis cs:1.025,-0.02);
\addplot[only marks, mark=*, mark size=1.6pt, draw=vAlibaba, fill=vAlibaba!45, forget plot] coordinates {(0.9980,-0.2060)};
\addplot[only marks, mark=square*, mark size=1.6pt, draw=vAlibaba, fill=vAlibaba!45, forget plot] coordinates {(1.0000,-0.0440)};
\addplot[only marks, mark=diamond*, mark size=1.6pt, draw=vAlibaba, fill=vAlibaba!45, forget plot] coordinates {(0.9654,-0.0808)};
\addplot[only marks, mark=triangle*, mark size=1.6pt, draw=vAlibaba, fill=vAlibaba!45, forget plot] coordinates {(0.9900,-0.0580)};
\addplot[only marks, mark=*, mark size=1.6pt, draw=vAlibaba, fill=vAlibaba!45, forget plot] coordinates {(0.9160,-0.1800)};
\addplot[only marks, mark=square*, mark size=1.6pt, draw=vAlibaba, fill=vAlibaba!45, forget plot] coordinates {(0.9900,-0.0300)};
\addplot[only marks, mark=triangle*, mark size=1.6pt, draw=vAlibaba, fill=vAlibaba!45, forget plot] coordinates {(0.9979,-0.0357)};
\addplot[only marks, mark=*, mark size=1.6pt, draw=vAlibaba, fill=vAlibaba!45, forget plot] coordinates {(0.9900,-0.1620)};
\addplot[only marks, mark=square*, mark size=1.6pt, draw=vAlibaba, fill=vAlibaba!45, forget plot] coordinates {(0.9900,-0.0380)};
\addplot[only marks, mark=triangle*, mark size=1.6pt, draw=vAlibaba, fill=white, line width=0.7pt, forget plot] coordinates {(0.7888,-0.1578)};
\addplot[only marks, mark=*, mark size=1.6pt, draw=vAlibaba, fill=white, line width=0.7pt, forget plot] coordinates {(0.6300,-0.0020)};
\addplot[only marks, mark=square*, mark size=1.6pt, draw=vAlibaba, fill=vAlibaba!45, forget plot] coordinates {(0.9000,-0.0120)};
\addplot[only marks, mark=diamond*, mark size=1.6pt, draw=vAlibaba, fill=vAlibaba!45, forget plot] coordinates {(0.9000,0.0000)};
\addplot[only marks, mark=triangle*, mark size=1.6pt, draw=vAlibaba, fill=vAlibaba!45, forget plot] coordinates {(0.9520,-0.0080)};
\addplot[only marks, mark=*, mark size=1.6pt, draw=vAlibaba, fill=white, line width=0.7pt, forget plot] coordinates {(0.6700,-0.0540)};
\addplot[only marks, mark=square*, mark size=1.6pt, draw=vAlibaba, fill=vAlibaba!45, forget plot] coordinates {(0.9080,-0.0200)};
\addplot[only marks, mark=diamond*, mark size=1.6pt, draw=vAlibaba, fill=vAlibaba!45, forget plot] coordinates {(0.9231,-0.0231)};
\addplot[only marks, mark=triangle*, mark size=1.6pt, draw=vAlibaba, fill=vAlibaba!45, forget plot] coordinates {(0.9620,-0.0060)};
\addplot[only marks, mark=*, mark size=1.6pt, draw=vAlibaba, fill=white, line width=0.7pt, forget plot] coordinates {(0.7420,-0.0220)};
\addplot[only marks, mark=square*, mark size=1.6pt, draw=vAlibaba, fill=vAlibaba!45, forget plot] coordinates {(0.9220,-0.0180)};
\addplot[only marks, mark=diamond*, mark size=1.6pt, draw=vAlibaba, fill=white, line width=0.7pt, forget plot] coordinates {(0.8962,0.0192)};
\addplot[only marks, mark=triangle*, mark size=1.6pt, draw=vAlibaba, fill=vAlibaba!45, forget plot] coordinates {(0.9680,0.0080)};
\addplot[only marks, mark=*, mark size=1.6pt, draw=vAlibaba, fill=vAlibaba!45, forget plot] coordinates {(0.9020,-0.0620)};
\addplot[only marks, mark=square*, mark size=1.6pt, draw=vAlibaba, fill=vAlibaba!45, forget plot] coordinates {(0.9920,-0.0160)};
\addplot[only marks, mark=diamond*, mark size=1.6pt, draw=vAlibaba, fill=vAlibaba!45, forget plot] coordinates {(0.9885,-0.0192)};
\addplot[only marks, mark=triangle*, mark size=1.6pt, draw=vAlibaba, fill=vAlibaba!45, forget plot] coordinates {(0.9920,-0.0120)};
\addplot[only marks, mark=*, mark size=1.6pt, draw=vAlibaba, fill=vAlibaba!45, forget plot] coordinates {(0.9860,-0.2980)};
\addplot[only marks, mark=square*, mark size=1.6pt, draw=vAlibaba, fill=vAlibaba!45, forget plot] coordinates {(0.9960,-0.1000)};
\addplot[only marks, mark=diamond*, mark size=1.6pt, draw=vAlibaba, fill=vAlibaba!45, forget plot] coordinates {(0.9962,-0.0269)};
\addplot[only marks, mark=triangle*, mark size=1.6pt, draw=vAlibaba, fill=vAlibaba!45, forget plot] coordinates {(0.9500,-0.0580)};
\addplot[only marks, mark=*, mark size=1.6pt, draw=vAlibaba, fill=vAlibaba!45, forget plot] coordinates {(0.9580,-0.1100)};
\addplot[only marks, mark=square*, mark size=1.6pt, draw=vAlibaba, fill=vAlibaba!45, forget plot] coordinates {(0.9920,-0.0120)};
\addplot[only marks, mark=diamond*, mark size=1.6pt, draw=vAlibaba, fill=vAlibaba!45, forget plot] coordinates {(0.9962,-0.0038)};
\addplot[only marks, mark=triangle*, mark size=1.6pt, draw=vAlibaba, fill=vAlibaba!45, forget plot] coordinates {(0.9040,-0.0400)};
\addplot[only marks, mark=*, mark size=1.6pt, draw=vAlibaba, fill=white, line width=0.7pt, forget plot] coordinates {(0.6700,-0.0540)};
\addplot[only marks, mark=square*, mark size=1.6pt, draw=vAlibaba, fill=vAlibaba!45, forget plot] coordinates {(0.9100,-0.0300)};
\addplot[only marks, mark=diamond*, mark size=1.6pt, draw=vAlibaba, fill=white, line width=0.7pt, forget plot] coordinates {(0.8808,-0.0115)};
\addplot[only marks, mark=triangle*, mark size=1.6pt, draw=vAlibaba, fill=vAlibaba!45, forget plot] coordinates {(0.9700,0.0020)};
\addplot[only marks, mark=*, mark size=1.6pt, draw=vAlibaba, fill=vAlibaba!45, forget plot] coordinates {(0.9040,-0.0720)};
\addplot[only marks, mark=square*, mark size=1.6pt, draw=vAlibaba, fill=vAlibaba!45, forget plot] coordinates {(0.9920,-0.0440)};
\addplot[only marks, mark=diamond*, mark size=1.6pt, draw=vAlibaba, fill=vAlibaba!45, forget plot] coordinates {(0.9923,-0.0154)};
\addplot[only marks, mark=triangle*, mark size=1.6pt, draw=vAlibaba, fill=vAlibaba!45, forget plot] coordinates {(0.9940,-0.0140)};
\addplot[only marks, mark=*, mark size=1.6pt, draw=vAlibaba, fill=vAlibaba!45, forget plot] coordinates {(0.9920,-0.2200)};
\addplot[only marks, mark=square*, mark size=1.6pt, draw=vAlibaba, fill=vAlibaba!45, forget plot] coordinates {(1.0000,-0.0480)};
\addplot[only marks, mark=diamond*, mark size=1.6pt, draw=vAlibaba, fill=vAlibaba!45, forget plot] coordinates {(1.0000,-0.0154)};
\addplot[only marks, mark=triangle*, mark size=1.6pt, draw=vAlibaba, fill=vAlibaba!45, forget plot] coordinates {(0.9260,-0.0460)};
\addplot[only marks, mark=*, mark size=1.6pt, draw=vAlibaba, fill=vAlibaba!45, forget plot] coordinates {(0.9860,-0.2220)};
\addplot[only marks, mark=square*, mark size=1.6pt, draw=vAlibaba, fill=vAlibaba!45, forget plot] coordinates {(0.9960,-0.0800)};
\addplot[only marks, mark=diamond*, mark size=1.6pt, draw=vAlibaba, fill=vAlibaba!45, forget plot] coordinates {(0.9885,-0.0577)};
\addplot[only marks, mark=triangle*, mark size=1.6pt, draw=vAlibaba, fill=vAlibaba!45, forget plot] coordinates {(0.9980,-0.0220)};
\addplot[only marks, mark=*, mark size=1.6pt, draw=vAlibaba, fill=vAlibaba!45, forget plot] coordinates {(0.9740,-0.3300)};
\addplot[only marks, mark=square*, mark size=1.6pt, draw=vAlibaba, fill=vAlibaba!45, forget plot] coordinates {(0.9940,-0.1020)};
\addplot[only marks, mark=diamond*, mark size=1.6pt, draw=vAlibaba, fill=vAlibaba!45, forget plot] coordinates {(0.9385,-0.1308)};
\addplot[only marks, mark=triangle*, mark size=1.6pt, draw=vAlibaba, fill=vAlibaba!45, forget plot] coordinates {(0.9740,-0.0700)};
\addplot[only marks, mark=*, mark size=1.6pt, draw=vAlibaba, fill=vAlibaba!45, forget plot] coordinates {(1.0000,-0.3320)};
\addplot[only marks, mark=square*, mark size=1.6pt, draw=vAlibaba, fill=vAlibaba!45, forget plot] coordinates {(0.9980,-0.1140)};
\addplot[only marks, mark=diamond*, mark size=1.6pt, draw=vAlibaba, fill=vAlibaba!45, forget plot] coordinates {(0.9808,-0.0577)};
\addplot[only marks, mark=triangle*, mark size=1.6pt, draw=vAlibaba, fill=vAlibaba!45, forget plot] coordinates {(0.9920,-0.0720)};
\addplot[only marks, mark=*, mark size=1.6pt, draw=vAlibaba, fill=vAlibaba!45, forget plot] coordinates {(0.9440,-0.1240)};
\addplot[only marks, mark=square*, mark size=1.6pt, draw=vAlibaba, fill=vAlibaba!45, forget plot] coordinates {(0.9440,-0.1400)};
\addplot[only marks, mark=diamond*, mark size=1.6pt, draw=vAlibaba, fill=vAlibaba!45, forget plot] coordinates {(0.9346,-0.1115)};
\addplot[only marks, mark=triangle*, mark size=1.6pt, draw=vAlibaba, fill=vAlibaba!45, forget plot] coordinates {(0.9360,-0.0880)};
\addplot[only marks, mark=*, mark size=1.6pt, draw=vAlibaba, fill=vAlibaba!45, forget plot] coordinates {(0.9780,-0.0860)};
\addplot[only marks, mark=square*, mark size=1.6pt, draw=vAlibaba, fill=vAlibaba!45, forget plot] coordinates {(0.9800,-0.0920)};
\addplot[only marks, mark=diamond*, mark size=1.6pt, draw=vAlibaba, fill=vAlibaba!45, forget plot] coordinates {(0.9846,-0.0692)};
\addplot[only marks, mark=triangle*, mark size=1.6pt, draw=vAlibaba, fill=vAlibaba!45, forget plot] coordinates {(0.9760,-0.0600)};
\addplot[only marks, mark=*, mark size=1.6pt, draw=vAlibaba, fill=vAlibaba!45, forget plot] coordinates {(0.9300,-0.1260)};
\addplot[only marks, mark=square*, mark size=1.6pt, draw=vAlibaba, fill=vAlibaba!45, forget plot] coordinates {(0.9300,-0.1140)};
\addplot[only marks, mark=diamond*, mark size=1.6pt, draw=vAlibaba, fill=vAlibaba!45, forget plot] coordinates {(0.9231,-0.1000)};
\addplot[only marks, mark=triangle*, mark size=1.6pt, draw=vAlibaba, fill=vAlibaba!45, forget plot] coordinates {(0.9220,-0.1140)};
\addplot[only marks, mark=*, mark size=1.6pt, draw=vAlibaba, fill=vAlibaba!45, forget plot] coordinates {(0.9120,-0.1520)};
\addplot[only marks, mark=square*, mark size=1.6pt, draw=vAlibaba, fill=vAlibaba!45, forget plot] coordinates {(0.9700,-0.0380)};
\addplot[only marks, mark=diamond*, mark size=1.6pt, draw=vAlibaba, fill=vAlibaba!45, forget plot] coordinates {(0.9923,-0.0308)};
\addplot[only marks, mark=triangle*, mark size=1.6pt, draw=vAlibaba, fill=vAlibaba!45, forget plot] coordinates {(0.9860,0.0020)};
\addplot[only marks, mark=*, mark size=1.6pt, draw=vAlibaba, fill=white, line width=0.7pt, forget plot] coordinates {(0.7800,-0.2040)};
\addplot[only marks, mark=square*, mark size=1.6pt, draw=vAlibaba, fill=vAlibaba!45, forget plot] coordinates {(0.9300,-0.0420)};
\addplot[only marks, mark=diamond*, mark size=1.6pt, draw=vAlibaba, fill=vAlibaba!45, forget plot] coordinates {(0.9115,-0.0500)};
\addplot[only marks, mark=triangle*, mark size=1.6pt, draw=vAlibaba, fill=vAlibaba!45, forget plot] coordinates {(0.9600,-0.0040)};
\addplot[only marks, mark=*, mark size=1.6pt, draw=vAlibaba, fill=vAlibaba!45, forget plot] coordinates {(0.9760,-0.2720)};
\addplot[only marks, mark=square*, mark size=1.6pt, draw=vAlibaba, fill=vAlibaba!45, forget plot] coordinates {(0.9960,-0.0440)};
\addplot[only marks, mark=diamond*, mark size=1.6pt, draw=vAlibaba, fill=vAlibaba!45, forget plot] coordinates {(0.9962,-0.0269)};
\addplot[only marks, mark=triangle*, mark size=1.6pt, draw=vAlibaba, fill=vAlibaba!45, forget plot] coordinates {(0.9920,-0.0360)};
\addplot[only marks, mark=*, mark size=1.6pt, draw=vAlibaba, fill=white, line width=0.7pt, forget plot] coordinates {(0.7280,-0.1680)};
\addplot[only marks, mark=square*, mark size=1.6pt, draw=vAlibaba, fill=vAlibaba!45, forget plot] coordinates {(0.9120,-0.0600)};
\addplot[only marks, mark=diamond*, mark size=1.6pt, draw=vAlibaba, fill=white, line width=0.7pt, forget plot] coordinates {(0.8269,-0.0654)};
\addplot[only marks, mark=triangle*, mark size=1.6pt, draw=vAlibaba, fill=vAlibaba!45, forget plot] coordinates {(0.9080,-0.0320)};
\addplot[only marks, mark=*, mark size=1.6pt, draw=vAlibaba, fill=vAlibaba!45, forget plot] coordinates {(0.9360,-0.1480)};
\addplot[only marks, mark=square*, mark size=1.6pt, draw=vAlibaba, fill=vAlibaba!45, forget plot] coordinates {(0.9900,-0.0300)};
\addplot[only marks, mark=diamond*, mark size=1.6pt, draw=vAlibaba, fill=vAlibaba!45, forget plot] coordinates {(1.0000,-0.0154)};
\addplot[only marks, mark=triangle*, mark size=1.6pt, draw=vAlibaba, fill=vAlibaba!45, forget plot] coordinates {(0.9860,-0.0100)};
\addplot[only marks, mark=*, mark size=1.6pt, draw=vAlibaba, fill=vAlibaba!45, forget plot] coordinates {(0.9680,-0.2080)};
\addplot[only marks, mark=square*, mark size=1.6pt, draw=vAlibaba, fill=white, line width=0.7pt, forget plot] coordinates {(0.8520,-0.0600)};
\addplot[only marks, mark=diamond*, mark size=1.6pt, draw=vAlibaba, fill=vAlibaba!45, forget plot] coordinates {(0.9385,-0.0308)};
\addplot[only marks, mark=triangle*, mark size=1.6pt, draw=vAlibaba, fill=vAlibaba!45, forget plot] coordinates {(0.9800,-0.0240)};
\addplot[only marks, mark=*, mark size=1.6pt, draw=vAlibaba, fill=vAlibaba!45, forget plot] coordinates {(0.9880,-0.1280)};
\addplot[only marks, mark=square*, mark size=1.6pt, draw=vAlibaba, fill=vAlibaba!45, forget plot] coordinates {(1.0000,-0.0920)};
\addplot[only marks, mark=diamond*, mark size=1.6pt, draw=vAlibaba, fill=vAlibaba!45, forget plot] coordinates {(1.0000,-0.0308)};
\addplot[only marks, mark=triangle*, mark size=1.6pt, draw=vAlibaba, fill=vAlibaba!45, forget plot] coordinates {(0.9740,-0.0260)};
\addplot[only marks, mark=x, mark size=3.2pt, draw=vMeta, line width=1pt, forget plot] coordinates {(0.9100,-0.0100)};
\addplot[only marks, mark=x, mark size=3.2pt, draw=vMeta, line width=1pt, forget plot] coordinates {(0.6840,-0.1240)};
\addplot[only marks, mark=x, mark size=3.2pt, draw=vMeta, line width=1pt, forget plot] coordinates {(0.8290,-0.0290)};
\addplot[only marks, mark=x, mark size=3.2pt, draw=vMeta, line width=1pt, forget plot] coordinates {(0.8810,-0.0120)};
\draw[->, vMeta, line width=0.6pt, opacity=0.8] (axis cs:1.0000,-0.0440) -- (axis cs:0.9100,-0.0100);
\draw[->, vMeta, line width=0.6pt, opacity=0.8] (axis cs:0.9980,-0.2060) -- (axis cs:0.6840,-0.1240);
\draw[->, vMeta, line width=0.6pt, opacity=0.8] (axis cs:0.9900,-0.0580) -- (axis cs:0.8290,-0.0290);
\draw[->, vMeta, line width=0.6pt, opacity=0.8] (axis cs:0.9654,-0.0808) -- (axis cs:0.8810,-0.0120);
\addlegendimage{only marks, mark=*, mark size=1.6pt, draw=vAlibaba, fill=vAlibaba!45}
\addlegendentry{coherence}
\addlegendimage{only marks, mark=square*, mark size=1.6pt, draw=vAlibaba, fill=vAlibaba!45}
\addlegendentry{factuality}
\addlegendimage{only marks, mark=diamond*, mark size=1.6pt, draw=vAlibaba, fill=vAlibaba!45}
\addlegendentry{preference}
\addlegendimage{only marks, mark=triangle*, mark size=1.6pt, draw=vAlibaba, fill=vAlibaba!45}
\addlegendentry{relevance}
\addlegendimage{only marks, mark=*, mark size=1.6pt, draw=vAlibaba, fill=white, line width=0.7pt}
\addlegendentry{ceiling $<0.90$}
\addlegendimage{only marks, mark=x, mark size=3.2pt, draw=vMeta, line width=1pt}
\addlegendentry{agent harness}
\end{axis}
\end{tikzpicture}
\caption{\textbf{Repeat ceiling against paraphrase cost, with the
agent-harness cells overlaid.} Marker shape gives the task; hollow markers fall
left of the $0.90$ readability bar. Crosses are Haiku through the agent
harness; each arrow runs \emph{from} that judge and task measured through a
direct API call \emph{to} the same cell measured through the harness.
Takeaway: a low ceiling pulls $\Delta\mathrm{JSS}$ toward zero rather than away
from it, so uncontrolled decoding hides wording sensitivity instead of exposing
it, and the harness arrows move left and up for that reason.}
\label{fig:ceiling}
\end{figure}

The cell treated separately in Appendix~\ref{app:supporting} has a ceiling of
$0.789$, so that judge disagrees with itself on more than a fifth of
byte-identical prompts, and sits below even the discarded pilot that set the bar
(Section~\ref{sec:bar}). Where the ceiling is $0.990$ or above the subtraction
is nearly a no-op and $\Delta\mathrm{JSS} \approx \mathrm{JSS} - 1$. We collect the
repeat arm everywhere regardless, because a judge run at a production
temperature above zero has a lower ceiling throughout.

\begin{table}[!ht]
\centering
\small
\caption{\textbf{Pooled contrasts with judges clustered by vendor.} This is
the confirmatory test. The estimator is the unweighted mean of judge-level
$\Delta\mathrm{JSS}$, with a cluster-robust (CR1) standard error at the vendor
level and $t$ inference on $G-1$ degrees of freedom; $p_{\text{SESOI}}$ is
one-sided against the $0.02$ threshold, Holm-corrected over four
contrasts. Takeaway: treating sibling checkpoints as independent overstates
precision by $1.4$ to $1.6$ times, and two of the four tasks lose their claim
to a practically meaningful effect once they are not.}
\label{tab:vendor}
\begin{tabular}{lrrcrrrl}
\toprule
\rowcolor{headBand}
Task & Judges & Vendors & \multicolumn{1}{c}{\footnotesize\itshape effect} & Mean & SE (clustered) & $p_{\text{SESOI}}$ & Verdict \\
\midrule
coherence  & $19$ & $6$ & \dvis{-0.181} & $\mathbf{-0.181}$ & $0.026$ & $0.0008$ & \textcolor{cBad}{clears threshold} \\
factuality & $24$ & $8$ & \dvis{-0.057} & $\mathbf{-0.057}$ & $0.012$ & $0.0076$ & \textcolor{cBad}{clears threshold} \\
preference & $20$ & $7$ & \dvis{-0.045} & $-0.045$ & $0.012$ & $0.0416$ & \textcolor{chipGrey}{not established} \\
relevance  & $24$ & $8$ & \dvis{-0.034} & $-0.034$ & $0.010$ & $0.1055$ & \textcolor{chipGrey}{not established} \\
\bottomrule
\end{tabular}
\end{table}

The point-null test still rejects on all four tasks under the clustered
standard error ($p$ from $9.5\times10^{-4}$ to $1.3\times10^{-2}$). The
unclustered version treats \texttt{qwen3-8b}, \texttt{qwen3-14b} and
\texttt{qwen3-32b} as three independent pieces of evidence about prompt
sensitivity, and they are not.

\subsection{Effect magnitudes against the threshold}
\label{sec:results-sesoi}

We declared $|\Delta\mathrm{JSS}| < 0.02$ to be not practically meaningful.
That threshold was set after the first judge's results were in hand
(Section~\ref{sec:metrics}), so it is a declared standard and not a
pre-registered one. A reader should know which of our thresholds preceded the
data and which did not.

\begin{table}[!ht]
\centering
\small
\caption{\textbf{Effect magnitude against the threshold.} ``Inside
threshold'' counts cells whose estimate falls within $\pm0.02$;
``underpowered'' counts those whose minimum detectable effect at their own
support exceeds the effect they report. Takeaway: on three of the four tasks
roughly half the cells cannot detect an effect of their own size, which is why
the confirmatory claims are pooled and clustered (Table~\ref{tab:vendor})
rather than counted cell by cell.}
\label{tab:magnitude}
\begin{tabular}{lrcrrr}
\toprule
\rowcolor{headBand}
Task & Readable cells & \multicolumn{1}{c}{\footnotesize\itshape effect} & Pooled $\Delta\mathrm{JSS}$ & Inside threshold & Underpowered \\
\midrule
coherence  & $19$ & \dvis{-0.181} & $-0.181$ & $0$ & $0$ \\
factuality & $24$ & \dvis{-0.057} & $-0.057$ & $4$ & $10$ \\
preference & $20$ & \dvis{-0.045} & $-0.045$ & $6$ & $13$ \\
relevance  & $24$ & \dvis{-0.034} & $-0.034$ & $9$ & $13$ \\
\bottomrule
\end{tabular}
\end{table}

The earlier version of this work, which ran three judges from one vendor,
reported factuality as a null on this reading. Twenty-four judges do not
support that: only four of its cells fall inside the threshold
(Table~\ref{tab:magnitude}). Relevance is
now the task closest to it, with nine of twenty-four cells inside, and we do
not report relevance as a null either. Thirteen of its cells cannot detect an
effect of their own size at their own support, and a small estimate from an
underpowered cell is not evidence of a small effect. Relevance is where the
evidence for a practically meaningful cost is weakest; coherence is where it is
strongest, with no readable cell inside the threshold and none underpowered.

\subsection{Most of coherence's excess is the width of its scale}
\label{sec:results-binarised}

A judge choosing among five answers has more ways to disagree with itself than
one choosing between two, so the comparison against the binary tasks is
confounded with the width of the decision space. A reviewer of the first
version asked for this control, and it substantially supports the objection. We
collapse each rating to \emph{high} ($4$ or $5$) against \emph{low} ($1$, $2$
or $3$), on both arms and on the repeat, and recompute the endpoint over the
nineteen readable coherence cells.

\begin{table}[!ht]
\centering
\small
\caption{\textbf{Coherence with its scale collapsed to binary.} Ratings of $4$ or $5$ become \emph{high} and $1$ to $3$ become \emph{low}, on both arms and the repeat. Takeaway: nearly three-fifths of coherence's excess is the width of its decision space rather than the task, though it remains the largest effect of the four.}
\label{tab:binarised}
\begin{tabular}{lcr}
\toprule
\rowcolor{headBand}
Coherence, $19$ judges & \multicolumn{1}{c}{\footnotesize\itshape effect} & Mean $\Delta\mathrm{JSS}$ \\
\midrule
five-point scale & \dvis{-0.181} & $-0.181$ \\
binarised        & \dvis{-0.074} & $-0.074$ \\
\midrule
shrinkage of the pooled means   & & $59.0\%$ \\
mean of the per-judge shrinkages & & $58.6\%$ \\
\bottomrule
\end{tabular}
\end{table}

We therefore state the headline claim in two parts wherever it appears. As the
tasks are actually posed, coherence loses roughly three times what factuality
loses. Held at a common decision width, it loses about a third more. The
first number is what a practitioner scoring on a five-point scale should expect;
the second isolates the task from its label set. The first is the operational
quantity and the second the scientific one; reporting
only the first would overstate what is special about judging coherence.

\subsection{Sensitivity is larger on the ordinal task than on the others}
\label{sec:results-spread}

The two exceptions to the fourteen-of-sixteen tendency are
\texttt{magistral-small}, which loses $0.140$ on factuality against $0.124$ on
coherence, and \texttt{mistral-medium}, which loses $0.092$ against $0.086$.
Both margins are inside the width of their own intervals, so they are better
read as ties than as reversals. The ordering at the bottom has also changed:
factuality was the smallest effect in the earlier three-judge version, and
across twenty-five judges relevance is smaller, so that ordering is superseded
rather than refined.

\begin{table}[!ht]
\centering
\small
\caption{\textbf{Dispersion across judges against dispersion across tasks.} Within-judge spreads are over the sixteen judges whose four cells are all readable. Takeaway: the two are comparable, which is why the earlier three-judge claim that task dominates judge is withdrawn.}
\label{tab:spread}
\begin{tabular}{lrlr}
\toprule
\rowcolor{headBand}
Within task, across judges & spread & Within judge, across tasks & spread \\
\midrule
relevance  & $0.122$ & minimum (\texttt{mistral-small}) & $0.026$ \\
factuality & $0.128$ & median                           & $0.146$ \\
preference & $0.131$ & maximum (\texttt{llama3-8b})     & $\mathbf{0.274}$ \\
coherence  & $\mathbf{0.270}$ &                         &          \\
\bottomrule
\end{tabular}
\end{table}

That version also concluded that the task matters more than which model does
the scoring. At three judges from one vendor the within-task spreads were
$0.014$, $0.044$ and $0.122$, against within-judge spreads of $0.124$ to
$0.162$. That comparison measured the similarity of three models from one
family. Across six providers the two quantities are comparable, and only
two of the sixteen judges have a within-judge spread exceeding the largest
within-task spread. On coherence the between-judge spread is the largest
dispersion anywhere in this study. We withdraw the claim instead of restating it
in a weaker form.

Instability on coherence is milder than a flip between distant ratings. Across
the twenty-five coherence cells the quadratic-weighted $\kappa$ has a median of
$0.820$, against exact-match agreement from $0.560$ to $0.892$ with a median of
$0.760$. The gap is the share of disagreement that is a one-point move, and it
is the larger share: \texttt{gemma-4-31b} agrees exactly on $0.848$ of items
but scores $0.941$ once distance is weighted.

\subsection{Scale does not predict stability within a family}
\label{sec:results-scale}

Comparing judges across vendors confounds scale with training recipe, so a
reviewer of the first version asked for a same-family ladder. The
\texttt{qwen-3} family supplies one: three sizes, one provider, one generation,
run under the same matched budget.

\begin{table}[!ht]
\centering
\small
\caption{\textbf{The \texttt{qwen-3} size ladder.} Three sizes, one provider, one generation, one matched budget; dashes are cells whose repeat ceiling falls below $0.90$. Takeaway: a fourfold increase in parameters buys no consistent gain, and relevance reverses at the largest size.}
\label{tab:ladder}
\begin{tabular}{lrrrr}
\toprule
\rowcolor{headBand}
Judge & factuality & coherence & relevance & preference \\
\midrule
\texttt{qwen3-8b}  & \dnum{-0.060} & --- & \dnum{-0.032} & --- \\
\texttt{qwen3-14b} & \dnum{-0.042} & --- & \dnum{-0.004} & \dnum{-0.050} \\
\texttt{qwen3-32b} & \dnum{-0.044} & \dnum{-0.272} & \dnum{-0.036} & \dnum{-0.027} \\
\bottomrule
\end{tabular}
\end{table}

The ladder is not monotone in any column, and two cautions apply. It is
incomplete in an informative direction: the two smaller models have more cells
below the readability bar than the largest does, so the smaller a judge is, the
less often its own ceiling permits reading its sensitivity at all. A second
intended ladder, the \texttt{gpt-oss} 20B/120B pair, is absent because its
provider's free tier could not finish a cell inside the time we had; we name it
instead of leaving the gap unexplained. The results establish something
narrower than a scaling law: within one family, parameter count did not
predict stability in our suite.

\subsection{The tasks discriminate, and coherence only narrowly}
\label{sec:results-discriminate}

Section~\ref{sec:shortcuts} established that no heuristic beats chance. The
complementary check requires the best judge on each task to exceed a threshold,
measured against the majority-class rate, not against uniform chance.

\begin{table}[!ht]
\centering
\small
\caption{\textbf{Discrimination against the majority-class rate.} A constant-answer judge already attains the majority-class rate, so quoting uniform chance would overstate every margin. Takeaway: three tasks discriminate comfortably and coherence does not, and coherence carries the largest paraphrase effect.}
\label{tab:discriminate}
\begin{tabular}{lrlrr}
\toprule
\rowcolor{headBand}
Task & threshold & best observed & majority class & judges clearing \\
\midrule
factuality & $0.75$ & $0.960$ (\texttt{gemini-3.7-flash}) & $0.500$ & \textcolor{okGreen}{$21/25$} \\
relevance  & $0.70$ & $0.899$ (\texttt{deepseek-v4-pro})  & $0.500$ & \textcolor{okGreen}{$25/25$} \\
preference & $0.65$ & $0.892$ (\texttt{gemma-4-31b})      & $0.500$ & \textcolor{okGreen}{$24/25$} \\
coherence  & $0.40$ & $0.452$ (\texttt{claude-sonnet})    & $0.348$ & \textcolor{cBad}{$6/25$} \\
\bottomrule
\end{tabular}
\end{table}

Two reviewers of the first version held that three of the four tasks failed to
discriminate. At twenty-five judges the opposite holds, and the task that
reverses is not the one they expected. Coherence's best judge exceeds a
constant-``4'' judge by ten points, and at $n=250$ the interval on $0.452$
reaches below the $0.40$ threshold, so coherence is only weakly discriminating.
That and its large paraphrase effect follow from the same property: a
five-point scale over gold labels that are a rounded three-annotator mean.

Answer-A rates confirm that the swap design holds against live judges. Across
the fifty pairwise cells the rate spans $0.045$ to $0.600$, with a median of
$0.518$ on relevance and $0.504$ on preference. A reviewer of the first version
reported that almost every judge returned label A on every pairwise item, which
saturated the metric; under the swap no cell saturates at the high end. The one
degenerate cell is \texttt{claude-sonnet} on preference at $0.045$, and it is
parse failure, not position behaviour: it carries a $43\%$ malformed
rate on the support it has, and it fell below the support floor. We leave it in
the range, because excluding the one cell that embarrasses the design would be
the wrong precedent.

\subsection{The same judge, run inside an agent harness}
\label{sec:results-harness}

Haiku was re-run on all four tasks through an agent harness that issues the
same items in batches of fifty and does not expose temperature.

\begin{table}[!ht]
\centering
\small
\caption{\textbf{The same judge through a direct API call and through an agent harness.} Model, items, templates and repeat design are identical; only the transport differs. Takeaway: both terms fall under the harness but the ceiling falls faster, so the measured sensitivity shrinks rather than grows.}
\label{tab:transport}
\begin{tabular}{llrrcr}
\toprule
\rowcolor{headBand}
Task & Transport & JSS & $\mathrm{JSS}_{\text{rep}}$ & \multicolumn{1}{c}{\footnotesize\itshape effect} & $\Delta\mathrm{JSS}$ \\
\midrule
factuality & direct API      & $0.956$ & $1.000$ & \dvis{-0.044} & $-0.044$ \\
           & agent harness   & $0.900$ & $0.910$ & \dvis{-0.010} & $-0.010$ \\
\addlinespace
coherence  & direct API      & $0.792$ & $0.998$ & \dvis{-0.206} & $-0.206$ \\
           & agent harness   & $0.560$ & $0.684$ & \dvis{-0.124} & $-0.124$ \\
\addlinespace
relevance  & direct API      & $0.930$ & $0.990$ & \dvis{-0.058} & $-0.058$ \\
           & agent harness   & $0.800$ & $0.829$ & \dvis{-0.029} & $-0.029$ \\
\addlinespace
preference & direct API      & $0.896$ & $0.965$ & \dvis{-0.081} & $-0.081$ \\
           & agent harness   & $0.869$ & $0.881$ & \dvis{-0.012} & $-0.012$ \\
\bottomrule
\end{tabular}
\end{table}

We report the harness as a second condition, not as a judge. Its cells
are excluded from Table~\ref{tab:main} and from every claim above, because they
carry no declared decoding configuration and are comparable only to themselves
across arms. Three alternative explanations were ruled out. All $102$ batches
returned their full complement of labelled items, no answer failed to parse,
and the repeat prompts were verified byte-identical to the arms they baseline.
One batch did come back eight items short; it was rejected and re-issued, since
a silently shortened batch would shift every later label within it.

The scope of the contrast is one model on four tasks. Only Haiku is reachable
under both transports, because the harness resolves its Sonnet and Opus aliases
to a later generation than the ones evaluated here. Two further judges were run
through the same harness on coherence and factuality, the tasks where the
collapse is largest and smallest. Neither has an API counterpart, so neither
supports a paired contrast.

\begin{table}[!ht]
\centering
\small
\caption{\textbf{Three judges through the same harness, on the task where the collapse is largest and the task where it is smallest.} Neither of the two additional judges has an API counterpart, so neither supports a paired contrast. Takeaway: the ceiling collapse is task-dependent, not model-dependent, and it is not ordered by model capability.}
\label{tab:harnessgrid}
\begin{tabular}{llrrcr}
\toprule
\rowcolor{headBand}
Task & Judge (agent harness) & JSS & $\mathrm{JSS}_{\text{rep}}$ & \multicolumn{1}{c}{\footnotesize\itshape effect} & $\Delta\mathrm{JSS}$ \\
\midrule
factuality & Haiku 4.5      & $0.900$ & $0.910$ & \dvis{-0.010} & $-0.010$ \\
           & Sonnet (later) & $0.980$ & $0.970$ & \dvis{+0.010} & $+0.010$ \\
           & Opus (later)   & $0.980$ & $0.990$ & \dvis{-0.010} & $-0.010$ \\
\addlinespace
coherence  & Haiku 4.5      & $0.560$ & $0.684$ & \dvis{-0.124} & $-0.124$ \\
           & Sonnet (later) & $0.520$ & $0.616$ & \dvis{-0.096} & $-0.096$ \\
           & Opus (later)   & $0.696$ & $0.848$ & \dvis{-0.152} & $-0.152$ \\
\bottomrule
\end{tabular}
\end{table}

A harness that simply injected decoding noise would have degraded both tasks.
Instead it appears to amplify borderline-ness the task already carries. This is
the reading two tasks and three models will bear, and not more: it does not
establish a mechanism, and three models from one vendor on two tasks cannot
settle the question for harnesses in general.

\subsection{Limits of this evidence}
\label{sec:results-limits}

Twenty-five judges across six providers is the breadth this design was built
for and could not report in its first version, but the roster is weighted
towards small and mid-sized checkpoints: three flagship models were dropped on
cost and four more were withdrawn when their providers' free tiers proved too
slow to finish a cell. More consequentially, \emph{it contains no OpenAI
model}. The GPT family was to be reached through a subscription harness, not the paid
API, and that route carries no declared decoding configuration, the same
objection that keeps our own agent-harness cells out of Table~\ref{tab:main}
(Section~\ref{sec:results-harness}). Mixing one vendor measured a different way
into a contrast that clusters by vendor would have been worse than leaving it
out. Every cross-vendor statement here is about the eight vendors in
Table~\ref{tab:vendor}, and the scale conclusions rest on two size ladders, not
on a sweep of the largest models available.

Two of the four tasks do not clear the threshold once judges are
clustered by vendor (Section~\ref{sec:results-sesoi}): preference falls just
outside after correction and relevance is not close, so on those two the
practical claim is unestablished, not merely small. Nor is the construct
assumption independently validated. No judge, human or model, scored the forty
instruction pairs for this release; the paper offers a deterministic structural
audit and the forty pairs in full (Section~\ref{sec:benchmark}).

Intervals cluster at the item, the right unit for the arms and repeats nested
inside an item, but items are themselves nested in source records, on every task
(Section~\ref{sec:sources}). The $250$ relevance items come from $50$
TREC-COVID topics, the $250$ coherence items from $92$ SummEval documents, the
$250$ factuality items from $125$ TruthfulQA records and the $130$ preference
items from $68$ MT-Bench questions. Every task therefore has an effective sample
smaller than its item count suggests, relevance most of all;
Appendix~\ref{sec:results-source-cluster} recomputes every pooled interval at
the record and no conclusion changes. The five templates are a fixed set, not a
sample, so every interval conditions on these particular strings and the
results describe sensitivity to \emph{these} rewordings.

Five cells are measured on terms the rest are not. The preference split is
complete for twenty-three of the twenty-five judges. Sonnet and Opus stopped at
$44$ and $170$ of $260$ rows when the API balance was exhausted mid-run, and
the pipeline refuses to emit $\Delta\mathrm{JSS}$ below the declared floor of
$100$ clusters, so those two cells report no endpoint rather than one computed
from $17$ and $80$ clusters. Separately, Opus~4.7 rejects an explicit
temperature parameter and ran at the provider default while the other
twenty-four judges ran at zero, so its three cells that do carry an endpoint
are not measured under the decoding policy the rest are; this is recorded on
every one of its call records, never inferred.

Every judge but that one is run at temperature zero, where the repeat ceiling
is highest. Production judges are often run above zero, or inside a harness that
fixes decoding on the caller's behalf. We expected the endpoint's separation to
grow under that noise; the one uncontrolled-decoding condition we have measured
(Section~\ref{sec:results-harness}) does the reverse, the ceiling falling
faster than paraphrase agreement so that $\Delta\mathrm{JSS}$ contracts and a
real effect becomes harder to detect.

Every prompt and every source corpus is English, so whether a judge stable in
English stays stable under translation-equivalent prompts is untested, and it
is a likely place for rankings to change. Every judge is also reached through a
single provider: alias drift, replica-level routing or an undocumented
checkpoint update would move every number in this paper at once, and we could
not detect it against an independent host. We record the provider-reported
model string on every call so such a change is visible after the fact.

\section{Conclusion}

We introduced \textsc{JudgeSense}, a benchmark of $880$ items asking whether a
judge returns the same verdict when the same request is worded differently,
scored against that judge's own agreement with itself. Rewording costs agreement on all four
tasks across twenty-five judges. On two of them the cost clears the threshold we
declare for a practically meaningful effect, once sibling checkpoints stop
counting as separate evidence, and the ordinal task loses $0.181$ where the
next loses $0.057$. We recommend that judge stability be reported
against a repeat baseline, measured in the transport the judge will run in, and
never read from a cell whose ceiling is low.


\ifblind\else
\subsubsection*{Acknowledgements}
Acknowledgements are omitted from the review copy and will be added for the
camera-ready.
\fi

\subsubsection*{Use of AI assistance}
Every quantitative claim in the paper is regenerated from committed artifacts
by a released script rather than transcribed, and the authors have verified them
against those artifacts. The authors take full responsibility for the content,
the analysis, and any remaining errors.

\subsubsection*{Data and code availability}
\label{sec:availability}
Everything needed to recompute the paper is released.

\begin{itemize}\itemsep2pt
\item \textbf{Benchmark.} The $880$ items across the four tasks, with both
  instruction templates per item, the position-swapped orderings for the two
  pairwise tasks, and the twenty templates in full.
\item \textbf{Raw judge outputs.} Every call record behind every number:
  $31{,}194$ rows over $100$ judge--task cells, each carrying the resolved
  decoding parameters and the provider-reported model string.
\item \textbf{Analysis.} The metric, the item- and source-clustered bootstraps,
  the multiplicity plan with its commit history, and the scripts that emit
  every table and both data figures in this paper.
\item \textbf{A standalone scorer} that computes the endpoint for a judge that
  is not in our roster, from a predictions file, with no API keys and no
  dependency on our provider clients.
\item \textbf{The heuristic baselines} that would defeat each task were the
  construction flawed, together with the scores they actually reach.
\end{itemize}

\noindent
\ifblind
Repository and dataset URLs are withheld for review; the anonymised artifact
accompanies the submission.
\else
Code: \url{https://github.com/rohithreddybc/judgeSense}. \
Dataset: \url{https://huggingface.co/datasets/Rohithreddybc/judgesense-benchmark}.
\fi

\subsubsection*{Reproducibility statement}
The artifacts listed above are released in full. Every table and both data
figures in Section~\ref{sec:results} are regenerated from
the committed raw outputs by a single script. Descriptive counts about the
construction, such as how many source comparisons each filter removed, are
reported from the loader's own logged output. Each split is content-hashed,
and a 32-check audit gate runs in continuous integration. Judge model
identifiers, resolved decoding parameters and the provider-reported model
string are recorded on every call record, so configuration drift is detectable
after the fact rather than assumed absent.

\appendix
\section{The polarity defect, and why no remapped analysis is reported}
\label{app:polarity}

The polarity constraint in Section~\ref{sec:benchmark} corrects a defect of our
own earlier design, which included a factuality template that inverted the
answer polarity (``does this response contain factual errors? answer NO if
accurate''). Under a single shared label space ``YES'' then denoted opposite
conclusions in different arms, so a judge answering both arms correctly was
recorded as disagreeing with itself. That is a label-convention defect in the
benchmark, not a property of any model, and it measures a different
construct: handling a polarity flip is a comprehension task, whereas we ask
whether equivalent phrasings yield equivalent verdicts. Excluding such pairings
biases the figures in a knowable direction, since polarity-inverted pairings
are among the hardest items in the set and removing them can only push measured
consistency upward.

The alternative is to retain inverted templates and remap each template's raw
answers onto one canonical decision space before comparison. That machinery is
implemented and released, but no remapped analysis is possible here, and not
because we are deferring the work: the v2 construction removes polarity
inversion at source, so all twenty shipped templates request the same answer
space in the same direction and there is nothing left for the remap to operate
on. Building a polarity-inclusive split would be a new release, and the
machinery is available for anyone who wants one.

The earlier version did quantify what the exclusion costs. Its magnitude rested
on data we have withdrawn, but the shape of the result stands. Including polarity-inverted pairings produced an apparent flip rate
near-identical across every judge tested, and a rate that does not vary with
the model is a property of the instrument, not of the models. That
uniformity is what identified the template as a label-convention defect rather
than a finding about judges. We keep the inference and discard the number.

\section{Discussion, construction detail and impact}
\label{app:discussion}

The body carries the measurement. This appendix carries what supports
it: what the results mean for a team choosing a judge, the shortcut and
outcome-category controls behind the construction, and the statement on
broader impact.

\subsection{Discussion}
\label{sec:discussion}

\subsubsection{What the result means for practice}

On the ordinal task, choosing a stronger judge does not reliably buy
stability, but neither does it fail to; the two moves are comparable in size.
Across the nineteen readable coherence cells the spread between the most and
least sensitive judge is $0.270$, the largest dispersion in this study and
wider than the median judge's spread across all four tasks ($0.146$;
Table~\ref{tab:spread}). A practitioner choosing a coherence judge is making a
decision that matters at least as much as the decision about response format,
which is the opposite of what the earlier version of this work concluded from
three judges of one vendor (Section~\ref{sec:results-spread}).

The obvious mechanism is the size of the answer space: a five-point scale gives
a rewording more ways to move the answer, and the weighted $\kappa$ shows most
of that movement is a single scale point. Collapsing the scale at $\geq 4$
removes $58.6\%$ of the coherence effect, averaged over judges
(Section~\ref{sec:results-binarised}). What survives still leads the four, so
the ordering holds, but the margin is about $1.3$ times factuality, not
three.

The same confound cuts the other way. A binary answer space quantises the
judge's latent uncertainty: a move from ``YES'' to a less confident ``YES'' is
recorded as perfectly consistent, while the same drift on a five-point scale
crosses a category boundary and is recorded as a flip. Relevance, the smallest
effect in the study at $-0.034$ and the task with the most cells inside the
declared threshold, is therefore consistent with two readings this design does
not separate: either judges really are stable when picking between two
passages, or a two-valued answer space cannot show the instability that is
there. A small binary effect is not licence to trust binary judges more.

\subsubsection{Refusal is an evaluation failure mode, not an aside}

One judge declined a quarter of the relevance items while another, on identical
prompts, declined none. For anyone using an LLM judge over medical or otherwise
sensitive corpora this is a first-order problem: the missing verdicts are not
missing at random, and a harness that silently treats a decline as a parse
failure will report a format-adherence problem that does not exist. A rewording
can also change whether a judge answers at all, which makes refusal an outcome
of the manipulation, so conditioning on the answered subset is selection on a
post-treatment variable. Worst-case bounds accompany the conditioned estimate
for that reason, and where refusals are common those bounds are wide.

\subsubsection{On building a benchmark that can be wrong}

The defects in our own rebuild are the most transferable part of this work.
Each passed every check that existed at the time. A deterministic
rotation assigning template pairs to items ran in lockstep with the alternating
emission of correct and incorrect answers, which made the template pair a
perfect answer key and skewed two templates' label balance to $75/25$, so a
constant-answer judge would have appeared to hold a large template preference.
Candidate responses reused across pairings carried contradictory gold labels,
so a judge reasoning correctly and consistently was scored wrong on nearly half
the preference items. A length-balancing step that correctly removed a
verbosity shortcut cut only one of the two buckets, and cut it by vote margin,
leaving the buckets matched on length and far apart on annotator agreement.

None of these is visible in labels, provenance, or aggregate statistics. All
three were found by asking what a heuristic with no understanding would score,
and by checking whether the artifact contained a rule that predicted the
answer. We would encourage that as routine: before reporting what models score
on a benchmark, report what nothing scores on it, with an interval and a
support count.

\subsubsection{Limitations}

Section~\ref{sec:results-limits} states what this evidence does not cover. Two
points bear repeating where the practical reading is made. The task ordering is
a strong tendency, not a rule: it holds for fourteen of the sixteen judges
whose four cells are all readable, and the two exceptions lose slightly more on
factuality than on coherence. And the effects are modest on three of the four
tasks. Only coherence and factuality clear the $0.02$ threshold once
judges are clustered by vendor; nine of twenty-four relevance cells and six of
twenty preference cells sit inside it, against four of twenty-four on
factuality, and thirty-six of the eighty-seven readable cells cannot detect an
effect of their own size. The ordinal result survives unambiguously. Coherence
loses roughly three times what the next task loses on its own scale, and about
a third more than factuality once that scale is collapsed.

\subsection{Broader impact statement}
The work is a measurement instrument for evaluation infrastructure, and its
risks are mostly the risks of being believed too readily.

A benchmark that reports stability invites the reading that a stable judge is a
good one. It is not: JSS never consults the ground-truth label, and a judge that
is wrong on every item but wrong identically scores perfectly. Selecting a judge
on stability alone would optimise for a model that is confidently and
reproducibly incorrect. Stability is necessary for an instrument and nowhere
near sufficient.

The refusal result has a distributional edge. One judge declined a quarter of
the relevance items, drawn from a medical corpus, while another on identical
prompts declined none. Where an evaluation harness silently drops declines, the
items lost are not lost at random, and the topics most likely to be dropped are
the sensitive ones. A pipeline that scores refusals as format failures will
report a clean evaluation over a corpus it partly refused to read. That failure
is quiet, and it gets quieter as the corpus gets more sensitive.

The benchmark is built from four existing English corpora and inherits whatever
they carry. Coherence gold is a rounded three-annotator mean, preference gold is
a human majority vote, and both encode the judgements of particular annotator
pools. Nothing here supports a claim about judge behaviour outside English, and
we have not measured it.

We see no dual-use concern specific to this artifact. It measures agreement
between two paraphrases of the same request; it does not generate content,
elicit capabilities, or make a model easier to misuse.

\subsection{Shortcut controls}
\label{sec:shortcuts}

We evaluate the shipped splits against judges that use no judgment at all, run
over the released files, not the loaders, so these are properties of the
artifact. Position only (always answer A) scores $0.500$ on relevance ($n=500$)
and $0.500$ on preference ($n=260$), by construction of the swap design. Length
only (pick the longer candidate) scores $0.482$ on relevance ($n=494$) and
$0.508$ on preference ($n=256$). Lexical overlap only (pick the candidate
sharing more terms with the query) scores $0.540$ on relevance ($n=252$).
Support counts fall below the row counts because tied rows are excluded rather
than split. All three come from \texttt{scripts/shortcut\_controls.py}, whose
output is committed.

None exceeds chance in either direction. The overlap control is the weakest:
its 95\% Wilson interval is $[0.478, 0.600]$, so the data are consistent with a
residual lexical signal worth about $0.57$, and it reads only half the
relevance rows because the remainder tie on the overlap statistic.

The length control is an enforced property of the item set. Annotators in the
preference corpus preferred the longer response in roughly 70\% of usable
comparisons, so we hold the winner-longer share at exactly 50\% by taking an
equal number of items from each length bucket. That $50\%$ is a construction
invariant. The $0.508$ above is what a length-following judge scores on the
rendered rows once ties are dropped.

The preference split ships 130 items, not the round 250: of the source
comparisons, 607 failed the decisive-vote label rule or carried no strict
majority, one was duplicate content, 67 had gold contradicting another item's
gold on shared candidate text, and 102 more went to hold the length balance
exactly. The last two filters were added after we found both failures in our
own shipped data.

Balancing on length alone would install a second confound, because only the
winner-longer bucket is cut and so any cutting criterion applies to one bucket
and not the other. Cutting by vote margin, as an earlier build did, left the
buckets matched on length but far apart on annotator agreement (mean
vote-margin ratio $0.567$ against $0.745$, $16\%$ against $54\%$ unanimous), so
a judge stronger on close calls would read as length-biased. We therefore
\emph{stratify}: the winner-shorter bucket fixes the target distribution of
vote shapes, absolute margin and total votes, and the winner-longer bucket is
filled stratum by stratum to match it. The shipped buckets agree to $0.0003$ in
mean vote-margin ratio ($0.8603$ against $0.8605$), have identical medians, and
contain $49$ unanimous comparisons each.

The opposite failure is equally real. The filter that removes contradictory
gold is a greedy selection over a conflict graph, so the order in which items
are considered determines which survive and how hard they are. Keeping the
firmest human majorities first is worse on both counts in this pool, giving
$106$ items at $90.6\%$ unanimous against $130$ items at $75.4\%$ when the most
contested are kept first, and a split on which annotators were nearly always
unanimous cannot separate a competent judge from a mediocre one. We resolve
conflicts contested-first.

The lexical control required a similar correction. Selecting each distractor to
maximise its term overlap with the topic made distractors \emph{denser} in
query terms than the relevant passages, so ``pick the passage with \emph{lower}
overlap'' recovered the answer far above chance. Matching the distractor's
retrieval score to the positive's removes overlap as a signal in either
direction, which is why the control is reported two-sided.

\subsection{Outcome categories: verdict, refusal, malformed}
\label{sec:refusal}

Each call terminates in exactly one of three outcomes. A \textbf{verdict} is a
response the strict parser maps to a label in the task's answer space. A
\textbf{refusal} is a response the provider itself flags as declined, typically
with empty content; providers spell this differently
(\texttt{stop\_reason="refusal"} for Anthropic,
\texttt{finish\_reason="content\_filter"} for OpenAI-compatible APIs,
\texttt{SAFETY} or \texttt{RECITATION} for Google), so the accepted set is
enumerated in one place in the released code, matched case-insensitively, with
the raw string retained on every record. Recognising only one family's spelling
would report a refusal rate of zero for every other family and silently charge
their declines as paraphrase disagreement. \textbf{Malformed output} is a
completed response the parser cannot map, recorded with its raw text and never
rounded to the nearest label.

The two non-verdict outcomes are treated differently. A malformed arm is
charged as disagreement and stays in the support, because the model completed a
response and the pair reflects a judgment the parser could not read. A refused
arm is removed from the support, because the provider halted before any
judgment was rendered: scoring a refusal as paraphrase disagreement asserts
that the judge produced two conflicting judgments, which it did not, and
treating refusal as a third label is worse still, since a judge that refuses
everything would then score $\mathrm{JSS} = 1.0$.

JSS is computed over pairs in which both arms returned verdicts and
report the support size $n$ alongside every figure. Refusal is reported per
cell as three quantities: the refusal rate (refused calls over all calls),
the refusal discordance rate (RDR; pairs in which exactly one arm was
refused), and the consistent refusal rate (pairs in which both arms were
refused). RDR is itself a sensitivity statistic. A nonzero RDR means a
meaning-preserving rewording changed whether the judge was willing to judge
at all, which is invisible to any analysis that scores refusals as
disagreement or discards them.

Conditioning on verdict pairs restricts a refusing judge to the items it agreed
to judge, so judges are compared on different subsets. We answer that with a
measurement: every cross-judge comparison in a task with nonzero refusals is
recomputed on the common support, the items every compared judge answered under
both phrasings, and beside it we report a refusal-inclusive figure in which
every refused arm counts as disagreement, the most punitive reading.

\section{Supporting analyses}
\label{app:supporting}

These are the robustness checks and single-cell diagnostics behind the
findings in Section~\ref{sec:findings}. They are reported in full here
instead of in the body, since a reader does not need them to follow the
argument, and because several answer requests made of the previous
version of this work.

\subsection{Testing against the threshold rather than against zero}
\label{sec:results-sesoi-test}

The contrasts above reject a null of no effect, and that null is weaker than it
looks. Under the strict policy $\Delta\mathrm{JSS}$ estimates a quantity that is
non-positive for any two prompts that do not induce identical response
distributions, so a point null of exactly zero is false before any data are
collected and rejecting it reports statistical power, not a property of
the judge. The test that carries the claim asks whether the effect is larger
than the smallest difference we declared of interest, and we report it in the
form a reader should hold us to: one-sided, $H_0: \overline{\Delta\mathrm{JSS}}
\geq -0.02$, Holm-corrected over the same four contrasts
(Table~\ref{tab:sesoi}).

\begin{table}[!ht]
\centering
\small
\caption{\textbf{Testing against the threshold rather than against zero,
treating judges as independent.} One-sided, $H_0:
\overline{\Delta\mathrm{JSS}} \geq -0.02$, Holm-corrected over four contrasts.
Table~\ref{tab:vendor} is the same test with judges clustered by vendor, and is
the confirmatory one. Takeaway: all four tasks clear the threshold here and
only two survive clustering, so the independence assumption is doing most of
the work.}
\label{tab:sesoi}
\begin{tabular}{lrcrrrl}
\toprule
\rowcolor{headBand}
Task & Judges & \multicolumn{1}{c}{\footnotesize\itshape effect} & Mean & Margin beyond $0.02$ & $z$ & $p$ (Holm) \\
\midrule
coherence  & $19$ & \dvis{-0.181} & $-0.181$ & $-0.161$ & $-8.47$ & $1.2\times10^{-17}$ \\
factuality & $24$ & \dvis{-0.057} & $-0.057$ & $-0.037$ & $-4.77$ & $9.4\times10^{-7}$ \\
preference & $20$ & \dvis{-0.045} & $-0.045$ & $-0.025$ & $-3.02$ & $1.3\times10^{-3}$ \\
relevance  & $24$ & \dvis{-0.034} & $-0.034$ & $-0.014$ & $-2.23$ & $1.3\times10^{-2}$ \\
\bottomrule
\end{tabular}
\end{table}

All four survive this version of the test, but the test assumes each judge is
an independent draw, and Section~\ref{sec:pooling} explains why they are not.
Under the vendor-clustered standard error we report as confirmatory
(Table~\ref{tab:vendor}), only coherence and factuality clear the
threshold. Preference falls just outside after correction and relevance is not
close. The ordering is the same either way, and the ordering is the point:
coherence clears by a wide margin under both, relevance clears under neither
once sibling checkpoints stop being counted separately. Had we set the
threshold at $0.05$ instead of $0.02$, only coherence would clear it.

\subsection{The conclusions survive clustering at the source record}
\label{sec:results-source-cluster}

Intervals elsewhere in this paper cluster at the item, the right unit for the
two arms and two repeats nested inside one item. It is not the only nesting
present, and all four tasks carry a second one. The $250$ relevance items are
drawn from $50$ TREC-COVID topics and the $250$ coherence items from $92$
SummEval documents; the $250$ factuality items come from $125$ TruthfulQA
records, each supplying one accurate and one inaccurate statement, and the $130$
preference items from $68$ MT-Bench questions. Two items sharing a record are
not independent draws, so item-clustered intervals are too narrow wherever that
dependence is strong. We reported this as a limitation of the first version, and
it is recomputable from the committed rows without issuing a single call.

Resampling source records instead of items, $2{,}000$ draws, gives
Table~\ref{tab:sourcecluster}.

\begin{table}[!ht]
\centering
\small
\caption{\textbf{Intervals clustered at the source record rather than the item.} Resampling records with replacement, $2{,}000$ draws. Takeaway: the nesting of items inside documents and topics changes no conclusion, so the item-clustered intervals reported elsewhere are not doing the work through an understated denominator.}
\label{tab:sourcecluster}
\begin{tabular}{lrlcrl}
\toprule
\rowcolor{headBand}
Task & Records & Unit & \multicolumn{1}{c}{\footnotesize\itshape effect} & Mean & 95\% CI (source-clustered) \\
\midrule
coherence  & $92$  & SummEval documents   & \dvis{-0.179} & $-0.179$ & $[-0.191, -0.167]$ \\
factuality & $125$ & TruthfulQA questions & \dvis{-0.058} & $-0.058$ & $[-0.067, -0.049]$ \\
preference & $68$  & MT-Bench questions   & \dvis{-0.052} & $-0.052$ & $[-0.066, -0.040]$ \\
relevance  & $50$  & TREC-COVID topics    & \dvis{-0.038} & $-0.038$ & $[-0.046, -0.030]$ \\
\bottomrule
\end{tabular}
\end{table}

Every interval excludes zero, and every interval excludes the $0.02$ threshold
as well. Relevance is where the nesting is heaviest, five items per topic on
average; factuality is where it is lightest, at two items per record. No pooled
mean moves by more than $0.001$ and no interval widens by more than $0.002$.

An earlier version of this analysis read the clustering unit off the item
naming scheme and reported factuality and preference as un-nested, at $250$ and
$130$ records. Both are nested, the dataset says so in a provenance field the
script did not consult, and for those two tasks what was printed as a
source-clustered interval was the item-clustered interval under another name.
The table above is computed from \texttt{source.source\_record\_id}, and the
script now fails rather than falling back to the item when a record identifier
does not parse.

We report this because the alternative is asking a reader to accept that a
known dependence does not matter. It does not matter here, and we measured that instead of assuming it.

\subsection{No judge wins by collapsing onto one label}
\label{sec:results-entropy}

A reviewer of the first version observed that our endpoint rewards compressed
output: a judge that answers ``YES'' to everything agrees with itself perfectly
under any rewording, and would score a perfect JSS for the worst possible
reason. The repeat ceiling does not catch this, because such a judge is
genuinely reproducible. The discrimination check in
Section~\ref{sec:results-discriminate} catches it only at the extreme. We
therefore record the decision distribution of every cell and report it here.

Table~\ref{tab:entropy} reports decision entropy, in bits, over the outcomes
each cell actually produced.

\begin{table}[!ht]
\centering
\small
\caption{\textbf{Decision entropy per task, in bits.} Computed over the
outcomes each cell actually produced, across twenty-five judges. The last
column is the entropy of a uniform distribution over the task's decision
space, which an observed value can exceed where a cell also produced refusals
or malformed output. Takeaway: no judge wins by collapsing onto a single label,
so the endpoint is not rewarding compressed output.}
\label{tab:entropy}
\begin{tabular}{lrrrr}
\toprule
\rowcolor{headBand}
Task & minimum & median & maximum & maximum possible \\
\midrule
factuality & $0.944$ & $1.000$ & $1.394$ & $1.000$ \\
coherence  & $1.031$ & $1.751$ & $2.076$ & $2.322$ \\
relevance  & $0.981$ & $1.052$ & $1.541$ & $1.000$ \\
preference & $0.881$ & $1.018$ & $1.393$ & $1.000$ \\
\bottomrule
\end{tabular}
\end{table}

No cell is close to collapse. The lowest entropy anywhere is $0.881$ bits
(\texttt{magistral-small} on preference), against $1.0$ for a judge splitting a
binary decision evenly; a judge answering one label on every item would score
$0$. On coherence, thirteen of the twenty-five judges use all five scale
points and the remaining twelve use four, so no judge treats the ordinal task
as binary.

Two of the maxima exceed the binary ceiling. Entropy is computed over the outcomes a cell
produced, and some cells produced an outcome outside the declared label set:
\texttt{claude-sonnet} returned \texttt{UNCLEAR} on four factuality arms, and
five judges emit it on coherence alongside the five ratings. Those are counted
as malformed in Section~\ref{sec:results-refusal} and never as a verdict, so
the endpoint is unaffected; the entropy column simply shows them.

\subsection{Refusal and malformed output are separate failures, and Sonnet has both}
\label{sec:results-refusal}

Sonnet declined $24.7\%$ of relevance arm calls, reported by the provider as a
refusal with empty content. Opus refused $5.1\%$ of the same prompts and Haiku
none at all. The corpus is TREC-COVID, so the items are medical, and the
effect is a property of one judge's safety behaviour, not of the
benchmark: the three judges saw identical prompts.

The outcome categories partition that cell's $1{,}000$ arm calls into $583$
verdicts, $249$ refusals and $168$ malformed responses. A harness that folded
refusals into malformed output would report a $41.7\%$ format-adherence
failure. The true malformed rate, computed over the arms the judge actually
attempted, is $22.4\%$. That is still a substantial and genuine
instruction-following failure, and the refusal finding should not be allowed
to obscure it.

Conditioning on the answered subset raises the fair objection that the three
judges are then scored on different items. We answer it by recomputing on the
common support, the 187 of 250 relevance items every judge answered under both
phrasings. The point estimates barely move: $-0.008$ for Haiku, $-0.001$ for
Opus and $+0.003$ for Sonnet. The conditioning is not what produces the
difference between these judges.

The deeper problem is not shared support. Because a rewording can itself
change whether a judge answers, refusal is one of the manipulation's outcomes
and not a nuisance to be dropped, and conditioning on the answered subset is
selection on a post-treatment variable. We report worst-case bounds. For this
cell the conditioned JSS is identified only within $[0.486, 0.746]$, a width
of $0.26$ over $130$ unidentified pairs. That region is wide, and we would
still rather show it than quote a point estimate resting on an assumption
nobody can check.

\subsection{Why Sonnet's relevance cell is not evidence about wording}
\label{sec:results-sonnet-relevance}

Three facts about that cell arrive together. Its repeat ceiling is $0.789$,
the lowest we observe. Its malformed rate differs by a factor of $2.7$ between
the two arms ($0.327$ against $0.122$). And its arm-specific repeat agreement
differs by $0.209$ ($0.685$ under one template, $0.893$ under the other).

That last figure is the diagnostic we built to detect templates that are not
exchangeable, and here it fires. If one arm fails to parse three times as
often as the other, the disagreement between arms is substantially a
difference in parse surface, not a difference in judgement, and
$\Delta\mathrm{JSS}=-0.158$ cannot be read as sensitivity to wording. The
paraphrase audit anticipated this: it warns that two of the relevance
templates state no single-token answer constraint, so their parse surfaces
differ from their partners'.

This cell is excluded from the ordering and spread claims above, and
report it as what it is: a judge whose behaviour on this corpus is dominated
by refusal and parse failure. It stays in Table~\ref{tab:main} and in the
count of intervals excluding zero, flagged rather than deleted.

\section{Experimental setup}
\label{app:setup}

Every judge below was reached through the provider named, at
\texttt{temperature}~$=0$ where the provider accepts it, with an identical
output budget of $1{,}024$ tokens and explicit suppression of reasoning traces
where the provider exposes that control. Whether suppression was requested and
whether it was honoured are recorded per call in the released logs, never
inferred. One judge rejects an explicit temperature parameter and therefore ran
at its provider default; its cells are marked in Table~\ref{tab:main}.

{\footnotesize
\begin{longtable}{llll}
\caption{\textbf{The twenty-five reported judges.} Alias as used throughout the
paper, the vendor that trained the checkpoint, the checkpoint identifier sent
to the provider, and the provider that served it. Vendor and provider differ
wherever a model is served by a host that did not train it. Takeaway: every
alias in Table~\ref{tab:main} resolves to a pinned checkpoint string.}
\label{tab:setup}\\
\toprule
\rowcolor{headBand}
Alias & Vendor & Checkpoint & Provider \\
\midrule
\endfirsthead
\multicolumn{4}{l}{\footnotesize\itshape Table~\ref{tab:setup}, continued.}\\
\toprule
\rowcolor{headBand}
Alias & Vendor & Checkpoint & Provider \\
\midrule
\endhead
\bottomrule
\endlastfoot
\texttt{claude-haiku} & Anthropic & \texttt{\scriptsize claude-haiku-4-5-20251001} & anthropic \\
\texttt{claude-opus-4-7} & Anthropic & \texttt{\scriptsize claude-opus-4-7} & anthropic \\
\texttt{claude-sonnet} & Anthropic & \texttt{\scriptsize claude-sonnet-4-5} & anthropic \\
\texttt{deepseek-v4-flash} & DeepSeek & \texttt{\scriptsize deepseek/deepseek-v4-flash} & novita \\
\texttt{deepseek-v4-flash-ds} & DeepSeek & \texttt{\scriptsize deepseek-v4-flash-0731} & dashscope \\
\texttt{deepseek-v4-pro} & DeepSeek & \texttt{\scriptsize deepseek-v4-pro-0813} & dashscope \\
\texttt{gemini-3.7-flash} & Google & \texttt{\scriptsize gemini-3.7-flash} & google \\
\texttt{gemini-flash} & Google & \texttt{\scriptsize gemini-2.5-flash} & google \\
\texttt{gemma-4-31b} & Google & \texttt{\scriptsize google/gemma-4-31B-it} & huggingface \\
\texttt{glm-5.2} & Zhipu & \texttt{\scriptsize glm-5.2} & dashscope \\
\texttt{kimi-k3} & Moonshot & \texttt{\scriptsize kimi-k3} & dashscope \\
\texttt{llama-3.3-70b} & Meta & \texttt{\scriptsize meta-llama/Llama-3.3-70B-Instruct} & huggingface \\
\texttt{llama-4-maverick} & Meta & \texttt{\scriptsize meta-llama/Llama-4-Maverick-17B-128E-Instruct-FP8} & huggingface \\
\texttt{llama-4-scout} & Meta & \texttt{\scriptsize meta-llama/Llama-4-Scout-17B-16E-Instruct} & huggingface \\
\texttt{llama3-8b} & Meta & \texttt{\scriptsize meta-llama/Llama-3.1-8B-Instruct} & huggingface \\
\texttt{magistral-small} & Mistral & \texttt{\scriptsize magistral-small-latest} & mistral \\
\texttt{mistral-medium} & Mistral & \texttt{\scriptsize mistral-medium-2604} & mistral \\
\texttt{mistral-small} & Mistral & \texttt{\scriptsize mistral-small-latest} & mistral \\
\texttt{qwen} & Alibaba & \texttt{\scriptsize qwen/qwen-2.5-72b-instruct} & novita \\
\texttt{qwen-3.6-flash} & Alibaba & \texttt{\scriptsize qwen3.6-35b-a3b} & dashscope \\
\texttt{qwen3-14b} & Alibaba & \texttt{\scriptsize Qwen/Qwen3-14B} & huggingface \\
\texttt{qwen3-32b} & Alibaba & \texttt{\scriptsize Qwen/Qwen3-32B} & huggingface \\
\texttt{qwen3-8b} & Alibaba & \texttt{\scriptsize Qwen/Qwen3-8B} & huggingface \\
\texttt{qwen3.7-flash} & Alibaba & \texttt{\scriptsize qwen3.7-flash-2026-07-15} & dashscope \\
\texttt{qwen3.8-27b-hf} & Alibaba & \texttt{\scriptsize Qwen/Qwen3.8-27B} & huggingface \\
\end{longtable}}

\paragraph{Judges registered but not reported.}
4 checkpoints are in the registry and carry no cell in
Table~\ref{tab:main}. None was dropped for its answers; each was withdrawn on
throughput, because the provider's free tier could not finish a cell in a
usable time. They are listed so the roster is auditable rather than curated.

\begin{center}
\begin{tabular}{lll}
\toprule
\rowcolor{headBand}
Alias & Vendor & Why it was withdrawn \\
\midrule
\texttt{gemini-3.1-pro} & Google & 250 requests/day, 16.3 days to finish \\
\texttt{gpt-oss-120b} & OpenAI & 200k tokens/day/model, 7.0 days to finish \\
\texttt{gpt-oss-20b} & OpenAI & 200k tokens/day/model, 5.3 days to finish \\
\texttt{qwen3.8-27b} & Alibaba & 200k tokens/day/model, 1.5 days to finish \\
\bottomrule
\end{tabular}
\end{center}

\section{The twenty instruction templates}
\label{app:templates}

Every template shipped with the benchmark is printed below, exactly as
the builder emits it. Placeholders in braces are filled per item; the
payload is otherwise unchanged between the two arms of a pair.

Each item is issued under exactly two of the five templates for its
task, assigned by a label-stratified seeded permutation over the ten
unordered pairs (Section~\ref{sec:benchmark}). No item appears under
more than one pair, so template identity is nested inside item identity
and no between-template contrast is identifiable from this design.

The audit described in Section~\ref{sec:benchmark} is run against these
strings: it checks that the requested answer set is identical within
each pair, that no template inverts polarity, that each names its own
task's construct, and that no pair is a near-duplicate. It cannot establish
that a competent reader maps two wordings onto the same question, so the
equivalence claim is stated as an assumption
and not as a result.

\subsection{Factuality (TruthfulQA)}
Requested answer set: \{YES, NO\} for all five templates.

\paragraph{T1.}
{\ttfamily\small\raggedright\sloppy
  Is this statement factually correct? Answer YES or NO only. \{text\}
\par}

\paragraph{T2.}
{\ttfamily\small\raggedright\sloppy
  Fact-check the statement below. Reply YES (correct) or NO (incorrect). \{text\}
\par}

\paragraph{T3.}
{\ttfamily\small\raggedright\sloppy
  Is the following statement accurate? Answer with one word: YES or NO. \{text\}
\par}

\paragraph{T4.}
{\ttfamily\small\raggedright\sloppy
  Evaluate the factual accuracy of this statement. Respond YES if it is correct, NO if it is not. \{text\}
\par}

\paragraph{T5.}
{\ttfamily\small\raggedright\sloppy
  Judge whether the statement is factually correct. Your entire answer must be YES or NO. \{text\}
\par}

\subsection{Coherence (SummEval)}
Requested answer set: $\{1,\dots,5\}$ for all five templates.

\paragraph{T1.}
{\ttfamily\small\raggedright\sloppy
  Rate the coherence of this text from 1 (incoherent) to 5 (very coherent). Reply with one number only. \{text\}
\par}

\paragraph{T2.}
{\ttfamily\small\raggedright\sloppy
  How coherent is the following text? Score it 1-5 (1 = incoherent, 5 = very coherent). Number only. \{text\}
\par}

\paragraph{T3.}
{\ttfamily\small\raggedright\sloppy
  Assign a coherence rating from 1 to 5, where 1 means incoherent and 5 means very coherent. Answer with a single digit. \{text\}
\par}

\paragraph{T4.}
{\ttfamily\small\raggedright\sloppy
  On a 1-5 coherence scale (1 = incoherent, 5 = very coherent), what score does this text deserve? Respond with the number only. \{text\}
\par}

\paragraph{T5.}
{\ttfamily\small\raggedright\sloppy
  Score this text's coherence: 1 (incoherent) through 5 (very coherent). Give only the number. \{text\}
\par}

\subsection{Relevance (BEIR TREC-COVID)}
Requested answer set: \{A, B\} for all five templates.

\paragraph{T1.}
{\ttfamily\small\raggedright\sloppy
  Which passage is more relevant to the query "\{query\}"? Answer A or B only. A: \{first\} B: \{second\}
\par}

\paragraph{T2.}
{\ttfamily\small\raggedright\sloppy
  For the query "\{query\}", which passage is more relevant? Reply with A or B. A: \{first\} B: \{second\}
\par}

\paragraph{T3.}
{\ttfamily\small\raggedright\sloppy
  Query: "\{query\}". Choose the more relevant passage. Answer A or B. A: \{first\} B: \{second\}
\par}

\paragraph{T4.}
{\ttfamily\small\raggedright\sloppy
  Considering the query "\{query\}", decide which passage is more relevant. Respond A or B only. A: \{first\} B: \{second\}
\par}

\paragraph{T5.}
{\ttfamily\small\raggedright\sloppy
  Given the query "\{query\}", pick the passage with higher relevance. Your answer must be A or B. A: \{first\} B: \{second\}
\par}

\subsection{Preference (MT-Bench human judgments)}
Requested answer set: \{A, B\} for all five templates.

\paragraph{T1.}
{\ttfamily\small\raggedright\sloppy
  Which response to "\{query\}" is better? Answer A or B only. A: \{first\} B: \{second\}
\par}

\paragraph{T2.}
{\ttfamily\small\raggedright\sloppy
  For the question "\{query\}", which response is of higher quality? Reply with A or B. A: \{first\} B: \{second\}
\par}

\paragraph{T3.}
{\ttfamily\small\raggedright\sloppy
  Question: "\{query\}". Choose the better response. Answer A or B. A: \{first\} B: \{second\}
\par}

\paragraph{T4.}
{\ttfamily\small\raggedright\sloppy
  Considering the question "\{query\}", decide which response is better. Respond A or B only. A: \{first\} B: \{second\}
\par}

\paragraph{T5.}
{\ttfamily\small\raggedright\sloppy
  Given the question "\{query\}", pick the stronger response. Your answer must be A or B. A: \{first\} B: \{second\}
\par}

\bibliographystyle{plainnat}          
\bibliography{references}

\end{document}